\documentclass[11pt]{article}

\usepackage[final]{acl}
\usepackage{subcaption}

\usepackage{times}
\usepackage{latexsym}

\usepackage[T1]{fontenc}

\usepackage[utf8]{inputenc}

\usepackage{microtype}

\usepackage{inconsolata}

\usepackage{graphicx}

\usepackage{microtype}
\usepackage{graphicx}
\usepackage{subcaption}
\usepackage{booktabs} 

\usepackage{hyperref}
\usepackage{caption}
\usepackage{subcaption, multirow, array}
\usepackage{float, placeins}
\usepackage{dblfloatfix}

\usepackage{amsmath}
\usepackage{amssymb}
\usepackage{mathtools}
\usepackage{amsthm,bm}

\usepackage{soul}

\theoremstyle{plain}

\theoremstyle{definition}

\theoremstyle{remark}

\usepackage[textsize=tiny]{todonotes}
\title{DeepInsert: Early Layer Bypass for Efficient and Performant Multimodal Understanding}

\author{%
  Moulik Choraria$^{\dagger,\parallel,}$\thanks{Correspondence: moulikc2@illinois.edu; $^{\parallel}$Equal contribution; Code: https://github.com/MoulikChoraria/DeepInsert} \And
  Xinbo Wu$^{\dagger,\parallel}$ \And
  Akhil Bhimaraju$^{\dagger,\parallel}$ \\
  \AND
  Nitesh Sekhar$^{\S}$ \And
  Yue Wu$^{\Phi}$ \And
  Xu Zhang$^{\S}$ \And
  Prateek Singhal$^{\psi}$ \And
  Lav R. Varshney$^{\zeta}$ \\
  \AND
  \textnormal{$^{\dagger}$UIUC,
  $^{\S}$Amazon, $^{\Phi}$Capital One, $^{\psi}$Apple, $^{\zeta}$ Stony Brook University}
}

\begin{document}

\maketitle

\begin{abstract}
Hyperscaling of data and parameter count in LLMs is yielding diminishing improvement when weighed against training costs, underlining a growing need for more efficient finetuning and inference without sacrificing performance. This is especially so for multimodal language models (MLMs), where the overhead of processing multimodal tokens can limit their practical viability. 
Parallely, recent work has uncovered implicit cross-modal alignment in the deeper layers of large MLMs, deepening our understanding of how MLMs process and encode information. Motivated by this, and our observation that MLMs naturally defer most cross-modal token interactions to deeper layers of the model, we propose a simple modification. Instead of concatenation with the language prompt at the start, we insert multimodal tokens directly into the middle, allowing them to entirely bypass the early layers. Our results with diverse modalities, (i) LLaVA \& BLIP for vision, (ii) LTU for audio, and (iii) MoLCA for molecular data, and model sizes, starting from 350M to 13B parameters, indicate that our method reduces both training and inference costs, while at least preserving, if not surpassing the performance of existing baselines. 

\end{abstract}

\section{Introduction}
\label{Introduction}

\begin{figure*}
    \centering
    \begin{subfigure}{0.5\textwidth}
        \centering
        \includegraphics[width=0.8\textwidth]{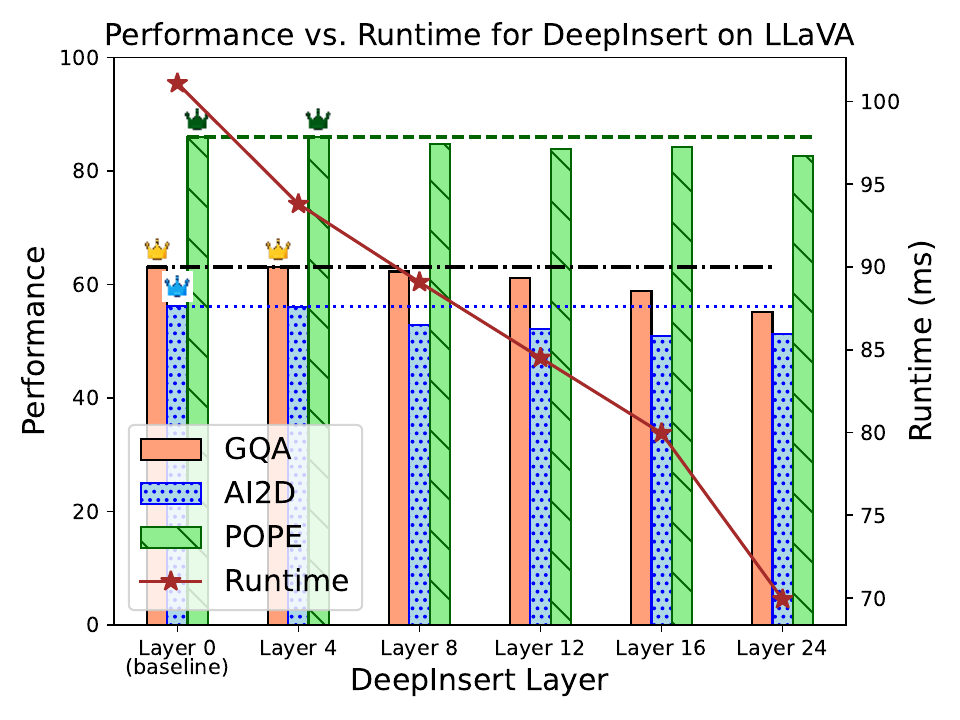}%
        \caption{Performance vs runtime in LLaVA (vision)}
        \label{fig:perf-vs-runtime-llava}
    \end{subfigure}%
    \begin{subfigure}{0.5\textwidth}
        \centering
        \includegraphics[width=0.8\textwidth]{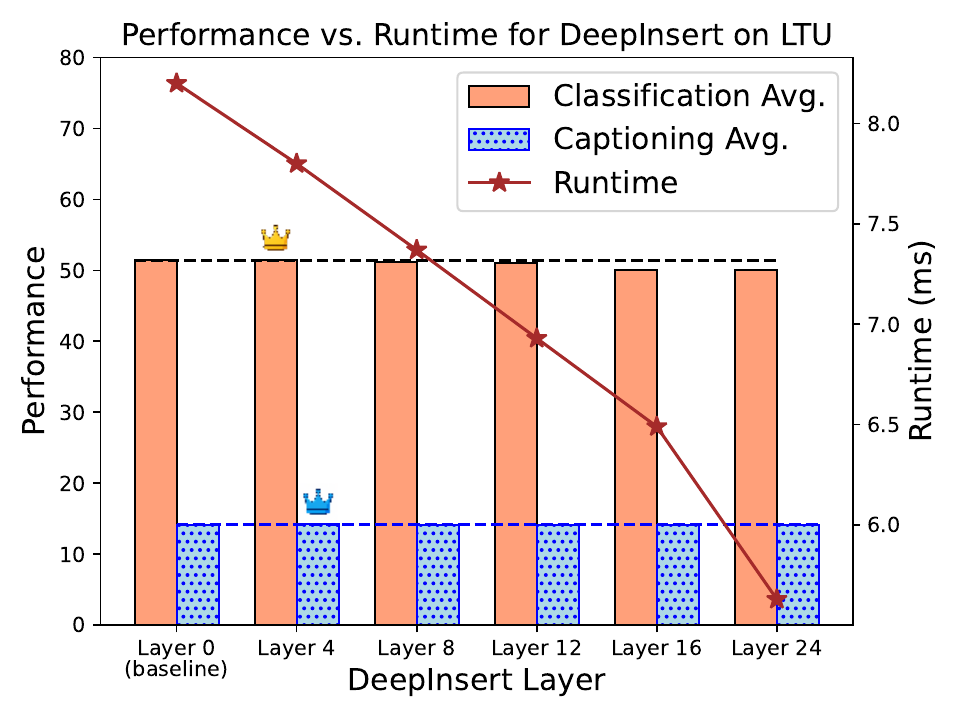}%
        \caption{Performance vs runtime in LTU (audio)}
        \label{fig:perf-vs-runtime-ltu}
    \end{subfigure}
    \caption{Tradeoff between performance and computational efficiency (inference) in (a) LLaVA v1.5-7B and (b) LTU-7B highlight our two contributions. First, we can reduce multimodal processing (efficiency gains), while either maintaining or improving performance with DeepInsert (see layer 4). Second, it is possible to stay competitive with the baseline, while having significant efficiency gains ($\sim 20$--$25\%$ speedup, see layer 12).}    \label{fig:tradeoff}
\end{figure*}




Advances in large language models (LLMs) \citep{brownMRSKA2020, chungHLZT_et_al_2022,touvron2023llama2openfoundation, openai2024gpt4} have spurred the development of models capable of multimodal comprehension in diverse application domains. While scaling has been crucial for this success \citep{kaplanMHBCCGRWA2020, zhaiKHB2021, hoffmannBMBCRC_etal_2022}, the computational costs associated with large multimodal language models (MLMs/MLLMs) \citep{strubellSGM2019, grattafioriDJPKDLM_etal_2024} necessitates methods for efficient finetuning and inference \citep{huSWALWWC2021, leviathanKM2023}. For multimodal tasks such as vision-language understanding  \citep{radfordKHKS2021, liLXH2022, liuLWL2024}, the use of pretrained unimodal models, rather than training from scratch, is effective at reducing training overhead \citep{qiLSWL2024}. However, overall efficiency remains a challenge, as multimodal inputs impose additional computational costs, when processed alongside language prompts. 

Another key to multimodal training is effective cross-modal alignment, which is achieved with a combination of pretraining stages, followed by multitask instruction finetuning for downstream tasks \citep{daiLLTZWLFH2023}. Preference tuning techniques can be used to produce further improvements \citep{ouyangWJAWMZA2022, sunSCLLSGGWYKD2023, wangCWZZYZGBHX2024}. However, most works in this area tend to propose modifications to the training algorithm and/or choice of dataset, while largely treating the pretrained components as replaceable black-box models. To complement this body of work from the efficiency point of view, we begin our investigation by analyzing the internal representations, relying on pretrained VLMs as a starting point.

Our initial inspiration comes from \citet{huhCWI2024}, who argue that representations from different unimodal models converge towards a universal platonic representation. Although the platonic hypothesis itself is contentious, the underlying evidence suggests that model representations across modalities show increasing alignment in the deeper layers (see Figs.~\ref{fig:alignment_main}, \ref{fig:alignment_dino} in the Appendix). This, coupled with the inner product mechanism of self-attention, invites the next question: in a multimodal setting, are cross-modal interactions more confined to deeper layers? The attention maps indeed suggest that the bulk of the activity is within the middle and latter layers (see Fig.~\ref{fig:attention_viz_avg_all}).
The preceding exposition culminates in our central investigation: \textit{Are the early MLLM layers necessary for multimodal token processing, and can we skip them without sacrificing performance?} More specifically,
\begin{enumerate}
    \item We demonstrate redundancy in early layers of multimodal LLMs, and that letting multimodal tokens bypass them \textbf{entirely} offers significant efficiency gains, with virtually no degradation. We refer to this framework as DeepInsert.  
    \item We showcase DeepInsert's versatility across modalities, via popular open-source MLLMs: vision (LLaVA, BLIP \citep{liuLLL2024, JunnanLSH2023}), audio (LTU \citep{gongLLKG2024}), and molecules (MolCA \citep{liuLLFCKWC2024}). We also reveal a natural performance-efficiency tradeoff (see Fig.~\ref{fig:tradeoff}), enabling practitioners to select models based on their specific needs. 
\end{enumerate}

The remainder of the paper is organized as follows. Sec.~\ref{sec:related_vlm} places our work in the context of broader literature. Sec.~\ref{sec:motivation_vlm} motivates our framework, while Sec.~\ref{sec:framework} details the general notion of the architectural modification, highlighting implementation and design considerations. Experiments in Sec.~\ref{sec:exp_vlm} demonstrate the versatility and effectiveness across a variety of popular open-source multimodal models. Sec.~\ref{sec:future_vlm} concludes with promising directions for broader application. 

\section{Related Work}
\label{sec:related_vlm}

\subsection{MLLMs \& Alignment}

The general architecture of MLLMs considered here comprises a pretrained LLM, frozen perceptual encoders for different modalities (image, audio, video, etc. \citep{gongCG2021, radfordKHKS2021, sunFWWC2023, zhaoGYZYS_etal_2024}), and a trainable mapping module (can be a lightweight MLP or a larger Perceiver \citep{jaegleGBZVC2021}) that aligns the modality with the LLM input space. The LLM prompts consist of text and multimodal tokens, the latter of which are obtained from the encoder via the mapping module. This general regime encompasses a wide range of works \citep{alayracDLBVZS2022, JunnanLSH2023, zhangLB2023, baiBCCDDFGHHH_etal_2023_qwen, liuLLL2024, gongLLKG2024, ghoshKSETSNDM2024, Microsoftphi2025}, and thus, DeepInsert is fairly universally applicable.

Based on where alignment is achieved, our method relates best to the idea of model stitching \citep{lencV2015, moschellaMFNLR2023}, that is, \textit{grafting} initial layers of a neural network onto the latter half of another with an affine layer. Our work can thus be interpreted as multimodal model stitching for efficiency, without any additional bells and whistles. 


\subsection{Efficiency via parameter reduction}

Low-rank adapters (LoRA) \citep{huSWALWWC2021, dettmersPHZ2023} are commonly used for computationally-efficient finetuning of LLM-based models.
The low-rank constraint enables finetuning large models with only a small set of additive parameters, reducing memory usage. However, inference remains just as costly. An alternative is model compression, where an existing model is effectively distilled, leading to accelerated inference and almost-at-par capabilities \cite{fangWHWWL2021, wangZZZ2022}. 
While these techniques exploit parametric redundancies in LLMs, our approach is motivated via functional redundancy, and therefore largely complementary.

\subsection{Layer skipping \& Token reduction}

Layer skipping, which has been employed in LLMs for speculative decoding and speeding up inference \citep{corroGAYAM2023, elhoushiSLHWL_etal_2024, dinKCG2024}, has been recently applied to MLLMs \citep{shukorC2024, ZengHJY2025}. Specifically, \citet{shukorC2024} combine layer skipping with compression and pruning of the base LLMs to demonstrate impressive efficiency gains in their custom setup. However, it is unclear if their method scales well as it falls $10-20\%$ short of baseline performance when trained in the LLaVA setup. 
On the other hand, \citet{ZengHJY2025} propose adaptive token skipping by selecting top-k representative tokens via cosine distances. However, their token selection is inherently LLaVA-centric, exploiting redundancies across vision tokens, and it is unclear if it generalizes to other VLMs or other modalities. FlexiDepth~\citep{luo2025flexidepth} and AdaSkip~\citep{luo2025adaskip} dynamically adjust the number of layers executed based on input complexity, while MoLe-VLA~\citep{zhang2025molevla} and \(\gamma\)-MoD~\citep{luo2024gammamod} introduce mixture-of-expert or modality-specific modules to selectively bypass computation. While effective within their respective domains, these approaches are often tailored to unimodal LLMs (FlexiDepth, AdaSkip) or require architectural modifications for multimodal settings (MoLe-VLA, \(\gamma\)-MoD). In contrast, DeepInsert, without relying on context-dependent routing or auxiliary modules, provides a principled framework for enabling full early layer skipping. Importantly, DeepInsert maintains strong performance across many modalities, while preserving the same training setup.

Another line of research focuses on reducing the number of input tokens processed during inference such as ST$^3$~\citep{zhuang2024st3}, Skip-Vision~\citep{ZengHJY2025}, FastV~\citep{chen2024fastv}, VTW~\citep{lin2024vtw}, PruMerge~\citep{shangCXLY2024}, MADTP~\citep{cao2024madtp}, and PACT~\citep{dhouib2025pact}, which prune redundant tokens or learn adaptive token selection strategies. 
DeepInsert is conceptually orthogonal to this direction: while these are either early-exit or token pruning techniques, 
DeepInsert is essentially a late-entry method, which enables it to be applied in conjunction to further improve efficiency. 
In fact, we \textit{demonstrate in Appendix~\ref{sec:token_reduction-integration} that DeepInsert, when combined with FastV, VTW or PruMerge, either maintains or improves than baseline}. This suggests that DeepInsert can make the model more robust to token pruning strategies. 


\section{Where Multimodal Tokens Matter}
\label{sec:motivation_vlm}

\begin{figure*}[tbp]
     \centering 
     \begin{subfigure}{0.47\textwidth}
         \includegraphics[width=\columnwidth]{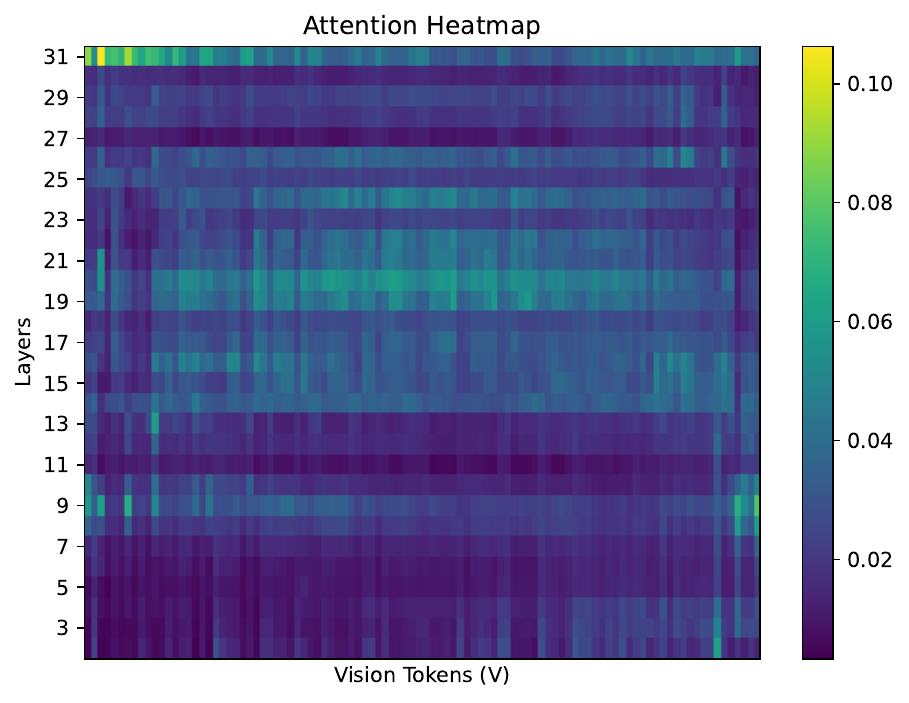}
         \caption{Vision to language token activation in LLaVA-7B. The y-axis represents LLM depth, the x-axis consists of vision tokens arranged in order of positional indices. The score represents the relative layerwise contribution of that token.  
         }
         \label{fig:attention_all_vision_tokens_separately}
     \end{subfigure}\hfill%
     \begin{subfigure}{0.49\textwidth}
         \includegraphics[width=\columnwidth]{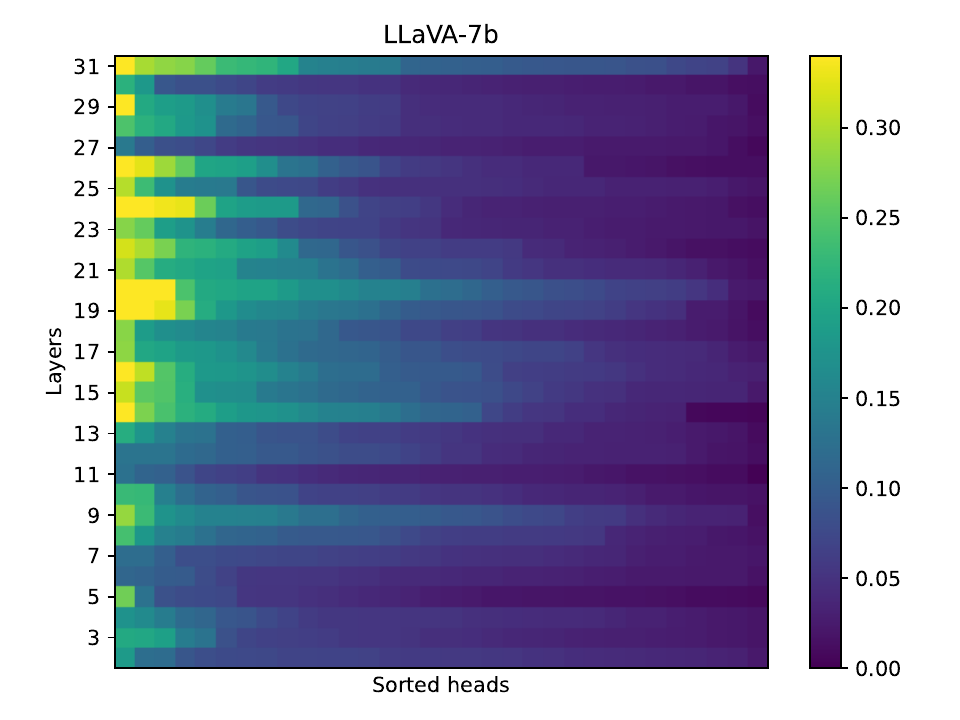}
         \caption{Visual Attention Ratio (VAR) \citep{jiangCZLSY2025} in LLaVA-7B. This is the sum of attention scores from all the vision tokens to the answer token, visualized as a function of attention heads (sorted left to right in decreasing order of activity).}
         \label{fig:var}
     \end{subfigure}
     \caption{Attention activity of vision tokens, visualized pre-prediction in Fig.~\ref{fig:attention_all_vision_tokens_separately} and post-prediction,
     in Fig.~\ref{fig:var}, jointly indicate that vision tokens become relevant only after the initial few layers.}
        \label{fig:attention_viz_avg_all}
\end{figure*}

This section motivates our proposed framework. We first explore the nature of representational alignment in pretrained unimodal models, in the setup of \citet{huhCWI2024}. While Figs.  \ref{fig:alignment_main}, \ref{fig:alignment_dino} in Appendix \ref{sec:VL_alignment} offer evidence of emergent semantic alignment in deeper layers, it is hard to quantify its role in multimodal training. Consequently, we turn towards analyzing the self-attention maps, to shed light on how the multimodal tokens engage with the language prompt during multimodal processing. We start with a balanced subset of multiple choice questions (MCQs) drawn from the Visual7W dataset \citep{zhuGBF2016}, based on COCO images \citep{chenFLVGDZ2015}. Using MCQ-based questions allows us to first isolate data samples where the model succeeds in answering correctly by comparing the token logit against the choices (A, B, C, or D) and then, obtain the relevant attention activity for these examples, in just one forward pass. Isolating the correct instances is important, since instances of hallucination can significantly alter the attention behavior \citep{jiangCZLSY2025}. 

With this setup in place, we analyze the attention activity as a function of model depth in two ways. (i) Fig.~\ref{fig:attention_all_vision_tokens_separately} shows the corresponding attention scores from the visual token to the last language token responsible for prediction. We subsample vision tokens (100 out of 576 tokens arranged left to right in order of positional indices)  and study their layer-relative contribution, by normalizing their per-layer contribution by its net contribution across layers. 
This allows us to cleanly visualize tokens that contribute high absolute value of attention scores (such as CLIP summary tokens \citep{neoOTGKB2025}) with those that do not. The attention scores correspond to the average of top-$k$ ($k=5$ heads for each layer) attention heads, with other cases in the Appendix \ref{sec:WVM_rem}.
(ii) Fig.~\ref{fig:var} shows the corresponding scores with respect to the first response/answer token post prediction. Inspired by \citep{jiangCZLSY2025}, we sum up attention scores from vision tokens to the answer token across all heads. Since the total incoming attention into a given token (per head) must sum to one, we obtain a ratio of vision to language attention in each layer. If this ratio is high, it implies the answer token receives most of the attention from vision tokens, rather than language tokens. Thus this is referred to as Visual Attention Ratio (VAR) \citep{jiangCZLSY2025}. We note a consistent trend in both cases: \textbf{vision tokens interact with language predominantly in intermediate layers}, indicating earlier layers may be redundant in multimodal processing.   


\textbf{Remark:} In the complete layer-wise depiction (Fig.~\ref{fig:attention_viz_avg_all_layers}), there is a strong activation pattern in the first two layers. 
However, this is unlikely to be the key to vision-to-language information transfer.
\citet{basuGMNFM2024, jiangCZLSY2025} suggest that this is LLaVA-specific and not observed for other VLMs/encoders. 
Moreover, our experiments in Appendix~\ref{sec:layer_heuristic} suggest that even in LLaVA, these layers are largely redundant.
So we skip the first two layers in Fig.~\ref{fig:attention_viz_avg_all} to keep the visualization cleaner.      

\section{Proposed Framework}
\label{sec:framework}

\begin{figure*}[tbp]
     \centering
     \begin{subfigure}{0.4\textwidth}
         \centering
         \includegraphics[width=0.85\textwidth]{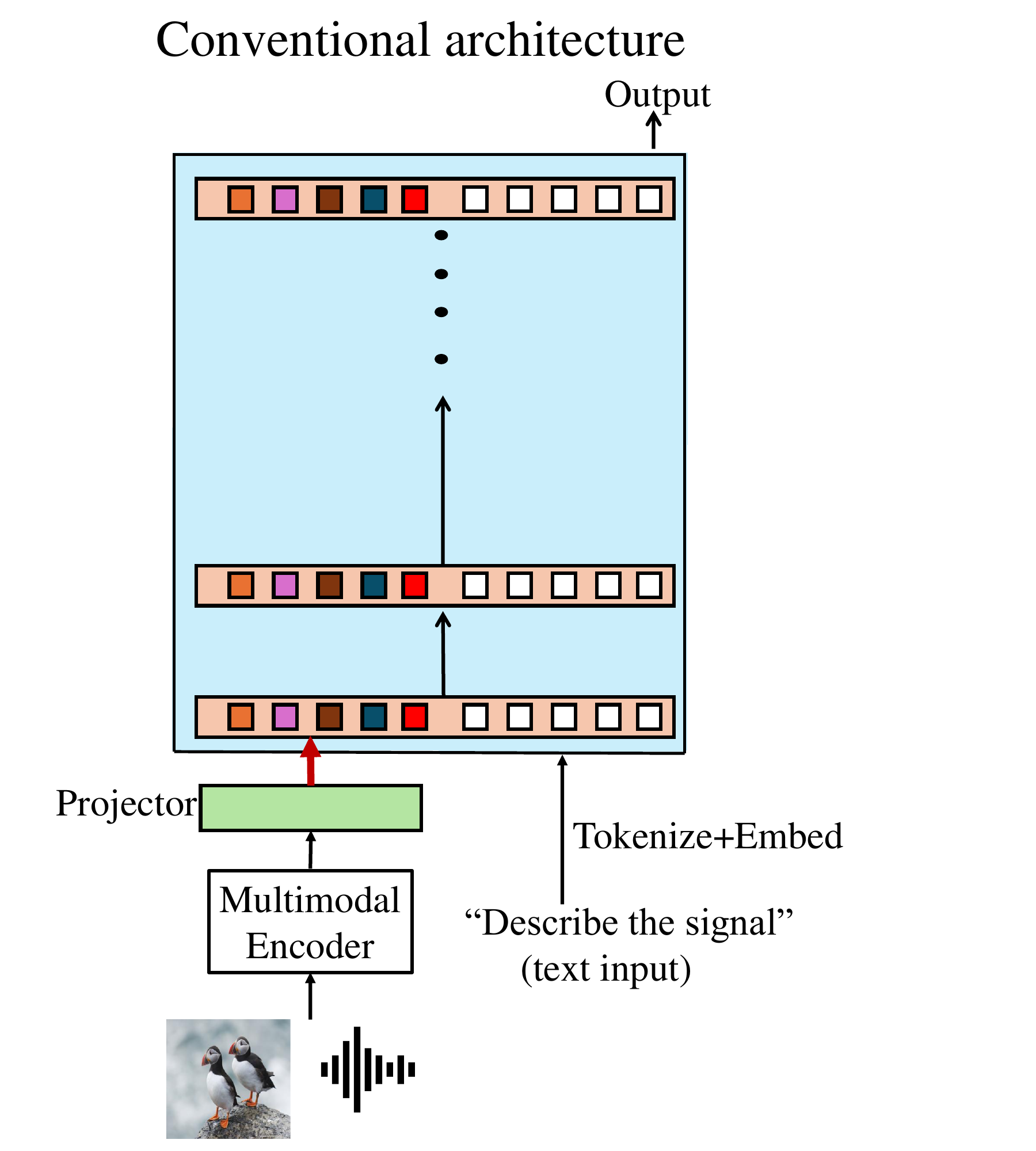}
         \caption{Conventional architecture for MLLMs.}
         \label{fig:image-conventional-architecture}
     \end{subfigure}%
     \begin{subfigure}{0.4\textwidth}
         \centering
         \includegraphics[width=0.85\textwidth]{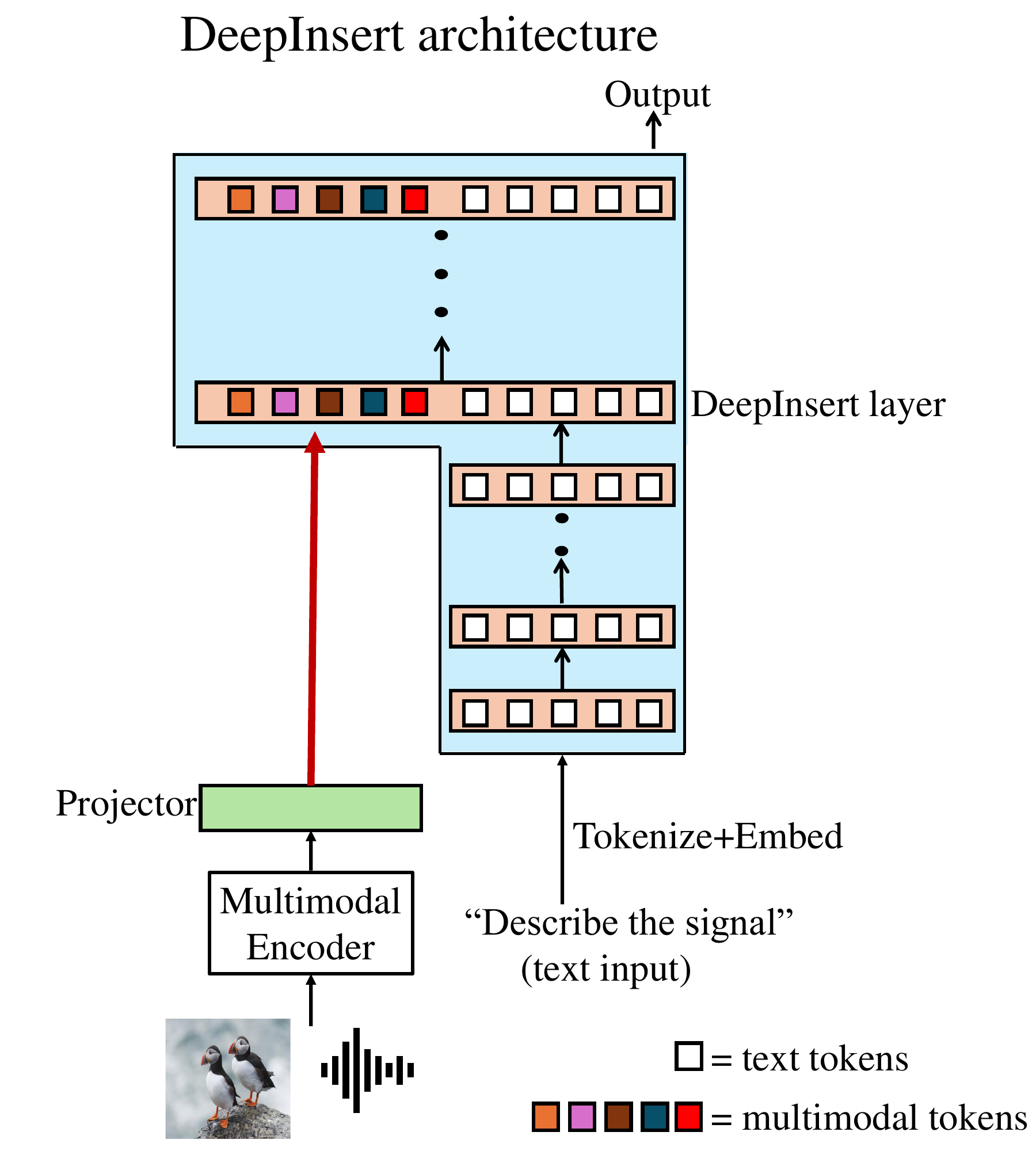}
         \caption{DeepInsert architecture for MLLMs.}
         \label{fig:image-deepinsert-architecture}
     \end{subfigure}
     \caption{The DeepInsert architecture contrasts with conventional MLLMs: We propose to entirely skip initial layers for the multimodal tokens to exploit underlying redundancies and improve computational efficiency.}
     \label{fig:architecture-comparison}
\end{figure*}

These observations bring us back to our central question: \textit{Do we really need the early LLM layers to parse multimodal information}? To investigate, we propose our simple yet effective modification for general MLLM frameworks, \emph{DeepInsert}: inserting the multimodal tokens directly into the intermediate layers of the LLM,
illustrated in Fig.~\ref{fig:architecture-comparison}.

\subsection{Technical Challenges}

Although straightforward as an abstraction, our approach poses several challenges. While multimodal tokens are inserted in the intermediate layer, language tokens must still be processed by the entire LLM for accurate language processing. Consequently, we must divide the prompt into language and multimodal constituents. The first stage of the forward pass processes just the language tokens. In the second stage, these are recombined before being processed by the latter layers of the LLM. Note that while this can be theoretically achieved by masking the vision tokens in the early layers, it naturally lowers efficiency gains and, we find, leads to noticeably worse performance. Therefore for our implementation, we need to refactor the LLM's forward pass as well as the KV-cache structure, so the model can maintain efficient generation capabilities. For interleaved text and multimodal data, we must also ensure consistency in positional embeddings through the splitting and recombining of the prompt.

\subsection{Design Considerations}

With implementation out of the way, the next question is: in which layer should the multimodal tokens be inserted? To prioritize efficiency, one can insert only in the last few layers. However, carelessly reducing the number of layers through which the tokens are processed will ultimately lower the capacity of the MLLM to exploit that modality, whether in moving information to language tokens or in memory recall within the MLP layers. At present, we sweep this tradeoff by training models with DeepInsert in different layers and comparing their performance. Indeed, characterizing the morphospace and associated Pareto frontier for the tradeoff between efficiency and learning capacity is an intriguing theoretical question. 

Nevertheless, woe befall us were we to let our reader depart empty-handed; thus, we firstly offer a compute-friendly heuristic for finding a suitable layer for DeepInsert. It turns out that by simply loading base MLLM weights onto the DeepInsert architecture and inserting multimodal tokens directly into the middle layers, one can achieve competent MLLMs. The corresponding performance degradations, as one inserts deeper, offer a good proportional signal for training the DeepInsert variant. We find this to be a nice and inexpensive sanity check for both LLaVA and LTU models (see Appendix \ref{sec:layer_heuristic}). With regards to the frontier, we extend this heuristic (see Appendix \ref{app:rl_layer_selection}) and offer a first step towards quantitatively characterizing this trade-off in a reinforcement learning setup, without needing to train the entire model from scratch.      

\section{Experiments}
\label{sec:exp_vlm}

We now test DeepInsert in a wide variety of settings across: a) different modalities (Vision: LLaVA \& BLIP), (Audio: LTU), and (Molecular graph data: MolCA), b) different base LLM sizes (ranging from 350M to 13B parameter models), and c) different LLM architectures (decoder-only \& encoder-decoder). To account for variations due to training settings (training setup, missing data, etc.), we also train the baseline versions for each modality. 
Experiments on the open-source models i.e.\ LLaVA, LTU, and MolCA are faithfully reproduced (see Sec.~\ref{sec:experiments-audio} for LTU missing data issues) as per the available training code. On the other hand, due to the scale as well as replicability issues with BLIP (unavailable code, hyperparameters and dataset sampling ratios, takedown of LAION dataset \citep{schuhmannBVGWCC_etal_2022, thiel2023}), we design a custom multitask training setup. For brevity, we move it to Appendix \ref{subsec:BLIP}. Finally, to demonstrate the applicability for newer, more multimodal intensive models, we present our results with the open reprduction of LLaVA-NeXT \citet{chenX2024} in Appendix \ref{subsec:llava_next}. 

For all our experiments with LLaVA, LTU, and MolCA, we use the default hyperparameter configurations from the respective repositories to stay fair to the baseline. While baselines for LLaVA and LTU replicate close to the reported results, MolCA shows noticeable performance gaps \citet{liuLLFCKWC2024}. However, both the baseline and the DeepInsert versions give us very similar performance, which verifies the essence of our claim. Most of the experiments (some training and all inference time estimates) were completed on Nvidia’s 80 GB A100 GPUs on a standalone machine. However, due to the scale of LLaVA-13B model and LTU training ($\sim 10m$ steps across all stages), we used 4 $\times$ 96 GB H100s (on an academic cluster), and 2$\times$ 141 GB H200 GPUs respectively. Finally, we will refer to insertion at layer X  as DeepInsert-X or X (DI) for brevity. 

\textbf{Remark.}
The choice of no additional hyperparameter tuning is deliberate, to emphasize that once we select an insertion layer, DeepInsert variants can easily match baseline performance and yield efficiency gains, while requiring no additional hyperparameter tuning when being adapted to open-source models in the wild.

\subsection{Vision}

LLaVA (Large Language-and-Vision Assistant) is an end-to-end multimodal model that combines a vision encoder with an LLM, along with a trainable projector MLP that maps visual features onto the LLM embedding space. The model is instruction-tuned in two stages: pretraining for feature alignment and finetuning for multimodal conversational tasks. 
Given its immense popularity and open-sourced training code/data, it represents a natural testbed for evaluating our framework at scale. 

We evaluate our method on LLaVA v1.5, with both 7B \& 13B models \citep{liuLLL2024} serving as our baseline. We use standard benchmarks for evaluation, including GQA \citep{hudson2019gqa}, TextVQA \citep{singh2019towards}, SciQA-IMG-IMG \citep{lu2022learn}, MME \citep{xu2024lvlm}, POPE \citep{li2023evaluating}, MMBench \citep{liu2024mmbench}, MM-Vet \citep{yu2023mm}, along with AI2D \citep{kembhavi2016diagram} to assess the model's ability to comprehend and interpret visual information within diagrams, and MMMU \citep{yue2024mmmu} to evaluate subject specific perception and reasoning with multimodal information. For a more detailed discussion of these benchmarks, the reader may refer to \citet{liuLLL2024}. 

Our results in Tables~\ref{tab:llava_performance},~\ref{tab:llava13b_performance} suggest that DeepInsert at layer 4 on average either matches or outperforms the baseline, while saving on compute. Beyond this point, the efficiency gains come at the cost of performance drops. DI-8 perhaps represents best value, with an average performance drop $\sim 1\%$, all while skipping $1/4$th (in 7B) and $1/5$th (13B) of the multimodal compute respectively. Note that MME is not included in the average calculation, so as to not skew the numbers.

\begin{table*}[t]
\begin{center}
\caption{Performance and runtime of LLaVAv1.5-7B models for various vision-language benchmarks. Bold and underline indicate the best and second-best performance, respectively. Our results indicate inherent redundancy in multimodal processing within LLaVA, which can be exploited by DeepInsert (DI-4), which can beat the baseline on average while offering efficiency gains.} 
\resizebox{\textwidth}{!}{
\begin{tabular} {@{}lccccccccc|cccc@{}}
 \toprule
 \textbf{\#Insert Layer} & GQA & 
 TextVQA & SciQA-IMG  & MME & POPE  & MMBench  & MM-Vet & AI2D & MMMU & Avg. &  Fwd. pass & Finetune \\
 & & & & & & & & & & (acc.) & (ms) & (hrs:min) \\
 \hline
 & & & & & & & & & & & \\
 0 (Baseline)  & \textbf{63.1} & \textbf{57.6} & \textbf{69.4} & 1436.2 & \textbf{86.0} & \textbf{66.1} & \underline{31.3} & \textbf{56.2} & \underline{36.1} & \underline{58.2} & 101.1 & 45:32 \\
4 (DI) & \textbf{63.1}  & \underline{57.5} & 68.7 & \textbf{1483.9} & \textbf{86.0} & \underline{66.0} & \textbf{32.6} & \underline{56.0} & \textbf{36.3} & \textbf{58.3} & 93.8 & 42:01\\
8 (DI) & \underline{62.3} & 57.1 & \underline{68.8} & \underline{1455.5} & \underline{84.8} & 65.3 & 28.8 & 52.9 & 34.9 & 56.9 & 89.1 & 39:46 \\
12 (DI) & 61.1 & 55.3 & 65.9 & 1248.5 & 83.9 & 55.2 & 30.3 & 52.2 & 35.6 & 54.9 & 84.5 & 36:48 \\
\bottomrule
\end{tabular}}
\label{tab:llava_performance}
\end{center}
\end{table*}

\begin{table*}[t]
\begin{center}
\caption{Performance of LLaVA v1.5-13B models for various vision–language benchmarks at different insert layers. We again verify that DeepInsert can maintain average performance, while offering efficiency gains. Note that we used a shared academic cluster for the training, where runtime and disk access speed can be affected by the usage patterns of other users. As a result, we may not observe a clear trend in the finetune times, as is the case here.}
\resizebox{\textwidth}{!}{
\begin{tabular}{@{}lccccccccc|cccc@{}}
\toprule
\textbf{\#Insert Layer} & GQA & TextVQA & SciQA-IMG & MME & POPE & MMBench & MM-Vet & AI2D & MMMU & Avg. & Fwd. pass & Finetune \\
 &  &  &  &  &  & &  &  & & (acc.) & (ms) & (hrs:min) \\
\midrule
0 (Baseline) 
& \textbf{63.7}
& \textbf{60.6} 
& \textbf{71.8} 
& \textbf{1593.6} 
& \textbf{87.1} 
& \underline{68.6} 
& \underline{38.3} 
& \underline{58.8} & \underline{35.3} & \textbf{60.5} & 154.4 & 26:55 \\
4 (DI) 
& \underline{63.5}
& \underline{60.0} 
& \underline{71.7} 
& \underline{1565.8} 
& \underline{87.0} 
& \textbf{69.0} 
& \underline{38.3} 
& \textbf{59.1} & 35.1 & \textbf{60.5} & 139.4 & 27:15 \\
6 (DI) 
& 63.0
& \underline{60.0} 
& 70.8 
& 1488.2 
& 86.7 
& 68.6 
& \textbf{38.9} 
& 58.7 & \textbf{36.7} & \underline{60.4} & 134.9 & 27:00 \\
8 (DI) 
& 63.0
& 59.6 
& 71.5 
& 1369.0 
& 86.8 
& 67.6 
& 36.0 
& 57.8 & 34.8 & 59.6 & 130.9 & 26:29 \\
\bottomrule
\end{tabular}}
\label{tab:llava13b_performance}
\end{center}
\end{table*}

\subsection{Audio}
\label{sec:experiments-audio}

LTU (Listen, Think and Understand), introduced by \citet{gongLLKG2024}, is a popular open-source Audio LLM. 
The model integrates an Audio Spectrogram Transformer \citep{gongCG2021}, finetuned as an Audio encoder, with a Vicuna-v1.5-7B \citep{vicuna2023} LLM to enable text-based outputs. The key contributor to its strong performance is the extensive pretraining and finetuning on the OpenAQA-5M dataset\footnote{As reported in \citet{ghoshKSETSNDM2024}, roughly $10\%$ of the data samples (mainly AudioSet \citep{gemmejeEFJLMPR2017}) are no longer usable during training.}, comprising 1.9 million closed-ended and 3.7 million open-ended audio-question-answer tuples, which helps facilitate perception and reasoning about audio. 

The training comprises 4 stages: stage-1 trains the audio projection layer with the closed-ended classification and acoustic feature description tasks. In stages 2--4, trainable LoRA adapters are introduced, and the complexity of the training task is gradually increased, including classification, acoustic feature description, and multiple closed-ended and open-ended audio language tasks. 

 As before, we consider LTU as our baseline, and compare with DeepInsert variants on benchmarks used by \citet{gongLLKG2024}, with five audio classification and two audio captioning tasks. 
 Classification accuracies are computed on the eval splits of the ESC50 \citep{piczak2015esc}, Vocal Sound \citep{gong2022vocalsound}, and VGG Sound \citep{chen2020vggsound} datasets. 
 Similarly, the mean average precision (mAP) classification score is computed on FSD50K \citep{fonseca2021fsd50k} and AudioSet \citep{gemmejeEFJLMPR2017}. Finally, SPICE scores \citep{anderson2016spice} are used to evaluate the models on captioning tasks using the AudioCaps \citep{kim2019audiocaps} and Clotho \citep{drossos2020clotho} datasets. The reader can refer to \citet{gongLLKG2024} for more details on evaluation, as we use the same process and metrics.

 Table~\ref{tab:ltu_performance} shows the performance of DeepInsert-LTU as a function of the insertion layer. We observe that insertion at layer~$4$ results in the best average performance for both classification and captioning tasks. Somewhat surprisingly, all models up to DI-24 stay fairly competitive with the baseline, as can be observed by the average performances, hinting at higher redundancies in audio.
 

 \begin{table*}
     \caption{Performance and runtime of LTU-7B for several audio-language benchmarks. 
     The best performance scores are in bold and the second best are underlined.
     We clearly see that models trained using multimodal insertion at deeper layers not only match baseline performance, but often exceed it.
     As expected, computational time for both training and inference drops as we insert deeper into the network
     (training was done on 2$\times$H200 141 GB GPUs).}
    \label{tab:ltu_performance}
    \begin{center}
    \resizebox{\textwidth}{!}{
        \begin{tabular}{lccccc|c|cc|c|cc}
            \toprule
             \textbf{\#Insert layer}& ESC50 & VS    & VGG   & FSD   & AudioSet & Classif. & AudioCaps & Clotho   & Cap. & Fwd. pass & Train time \\
                                    & (Acc) & (Acc) & (Acc) & (mAP) & (mAP)    &  Avg.    & (SPICE)   & (SPICE)  & Avg. &   (ms)    &  (hrs:min)\\
            \hline
            &&&&&&&&&&&\\
            0 (Baseline) & 85.45     & 56.42     & 49.64     & \ul{46.51}& 19.13     & \ul{51.43}& \ul{16.55}& 11.77	 & 14.16     & 8.20 & 17:30 \\
             4 (DI)      & \bf{86.10}& \bf{57.98}& 49.80     & 44.50     & 18.94	    & \bf{51.46}& 16.46	    & \bf{12.06}& \bf{14.26}& 7.80 & 16:16 \\
             8 (DI)      & 84.45     & \ul{57.34}& 48.87     & \bf{46.57}& 18.97     & 51.24     & 16.32     & \ul{12.01}& \ul{14.17}& 7.37 & 15:38 \\
             12 (DI)     & 85.10     & 56.42     & \ul{49.95}& 44.73     & \ul{19.23}& 51.09     & 16.37	    & 11.82	    & 14.10     & 6.93 & 15:04 \\
             16 (DI)     & \ul{85.90}& 52.74     & 49.64     & 42.93     & 19.15     & 50.07	    & \bf{16.62}& 11.71     & \ul{14.17}& 6.49 & 14:26 \\
             24 (DI)     & 82.30     & 53.41     & \bf{50.39}& 44.92     & \bf{19.51}& 50.11     & 16.30	    & 11.88     & 14.09     & 5.63 & 13:22 \\
            \bottomrule
        \end{tabular}     
    }
    \end{center}
 \end{table*}

\subsection{Molecular Data}

\begin{table*}[t]
    \centering
    \vspace{0.2cm}
     \captionsetup{labelformat=default} 
    \caption{Molecular captioning performances on PubChem324k and CheBI-20 datasets. Bold and underline indicate best and second-best performance, respectively. Both DeepInsert models (DI-9, DI-12), achieve similar or better performance than the baseline, using roughly only half the the LLM for multimodal processing.}
    \resizebox{0.7\textwidth}{!}{
    \begin{tabular}{@{}lcccccc@{}}
        \toprule
        \textbf{\#Insert Layer} & 
        \textbf{BLEU-2} & \textbf{BLEU-4} & \textbf{ROUGE-1} & \textbf{ROUGE-2} & \textbf{ROUGE-L} & \textbf{METEOR} \\ 
        \midrule
        \multicolumn{7}{l}{} \\ 
       0 (Baseline) & \underline{58.3} & \underline{49.5} & 66.1 & 51.4 & 59.9 & \underline{61.3} \\ 
       9 (DI) & \textbf{58.6} & \textbf{49.8} & \textbf{66.4} & \textbf{51.7} & \textbf{60.3} & \textbf{61.6} \\ 
       12 (DI) & {58.2} & \underline{49.5} & \underline{66.2} & \underline{51.6} & \underline{60.1} & \underline{61.3} \\  
        \bottomrule
        \vspace{0.2mm}
    \end{tabular}}\\
    
    {\parbox{0.9\textwidth}{\footnotesize
    \centering
    (a) Molecular Captioning on the CheBI-20 dataset for Galactica-1.3B based MolCA models.}}

    \vspace{0.4cm}

    \resizebox{0.7\textwidth}{!}{
    \begin{tabular}{@{}lcccccc@{}}
        \toprule
        \textbf{\#Insert Layer} & 
        \textbf{BLEU-2} & \textbf{BLEU-4} & \textbf{ROUGE-1} & \textbf{ROUGE-2} & \textbf{ROUGE-L} & \textbf{METEOR} \\ 
        \midrule
        \multicolumn{7}{l}{} \\ 
       0 (Baseline) & {34.1} & {26.4} & 48.5 & 33.9 & 43.1 & {41.8} \\ 
       9 (DI) & \underline{34.8} & \underline{27.2} & \underline{49.4} & \underline{34.7} & \underline{43.8} & \underline{43.0} \\ 
       12 (DI) & \textbf{35.3} & \textbf{27.7} & \textbf{49.7} & \textbf{35.3} & \textbf{44.3} & \textbf{43.4} \\  
        \bottomrule
        \vspace{0.2mm}
    \end{tabular}}\\
    {\parbox{0.9\textwidth}{\footnotesize
    \centering
    (b) Molecular Captioning on the Pubchem324k dataset for Galactica-1.3B based MolCA models.}}
   \label{tab:molca_best_performance}
\end{table*}

MolCA (Molecular Graph-Language Modeling with Cross-Modal Projector and Uni-Modal Adapter) enables language models to process both molecular graphs and text by bridging the two modalities \cite{liuLLFCKWC2024}. The training pipeline is almost identical to InstructBLIP: it uses a Cross-Modal SciBERT QFormer \cite{beltagyLC2019}, to translate molecular graph features into learnable query tokens that are fed into the LLM. Additionally, a unimodal low-rank adapter is used to fine-tune the LLM efficiently. The primary tasks include molecular-captioning and IUPAC name prediction. 

For our experiments, we follow the exact same training regime as \citet{liuLLFCKWC2024}; We initialize the model with the stage 1-pretrained QFormer, and start with the pretraining stage 2 for aligning molecular graphs with texts (10 epochs on Pubchem324k pretrain split \cite{liuLLFCKWC2024}), followed by finetuning unimodal adapters for downstream datasets. We focus on the molecular captioning experiments using Pubchem324k and CheBI-20 \cite{edwardsLRHCJ2022}, as the code for IUPAC task is not open-sourced. For the LLM, we consider the largest model, Galactica-1B \citep{taylorKCSHSPKS2022} with 24 layers, with identical LoRA configurations. DeepInsert injects the QFormer queries into different layers and we report performance via captioning evaluation metrics in \citet{liuLLFCKWC2024}. For both datasets, we train the model for the reported 100 epochs and check scores every 10 epochs. Since final performance is not the best for the baseline, we report the best model performance obtained during the entire training run. 

Results in Table \ref{tab:molca_best_performance} are consistent with our findings in other modalities; DI-9 and DI-12 can match and/or outperform the baseline model with ease, on both datasets. Due to the inherent drawbacks of captioning metrics used in the paper, it is harder to quantify the marginal gains as such. Notwithstanding, the result indicates the redundancy of multimodal processing also for molecular data, up to even $50\%$ (DI-12 ends up using only uses half the LLM layers for the multimodal tokens).

\subsection{Reflections on Performance vs. Efficiency}

Our results validate the general efficacy of DeepInsert in reducing multimodal redundancy. We also include training and inference run-times in Tables \ref{tab:llava_performance}, \ref{tab:llava13b_performance} and \ref{tab:ltu_performance} to demonstrate the efficiency of our implementation. This trade-off can be visualized as a function of insertion layer in Fig. \ref{fig:tradeoff}, where we train models for LLaVA and LTU with insertions upto layer 24, report their respective performances on a representative set of benchmarks and overlay it against the average inference forward pass time. Furthermore, Appendix \ref{sec:app_efficiency} gives a more objective breakdown of gains in terms of expected FLOPs, since above listed run-times are subjective due to varying GPU/compute capabilities.

A natural question here is why certain models and modalities (for instance, audio) demonstrate higher redundancy. Specifically, while both LTU and MolCA can be reduced to using only half the LLM with virtually no degradations, LLaVA performance seems to degrade faster after layer 8. We speculate that the major contributing factor here is down to the number of multimodal tokens. In general usage, while both LTU and MolCA compress modality information into 32 QFormer tokens, LLaVA uses all 576 vision tokens from the encoder. Consequently, this may lower redundancy due to the increase in multimodal load. Additionally, redundancy in  MolCA and LTU may be more due to higher training exposures (higher ratios of data/training iterations to the number of multimodal tokens, when compared to LLaVA).      

\section{Conclusion \& Future Work}
\label{sec:future_vlm}

In this work, we first analyze attention maps in multimodal models, revealing significant redundancy in their early layers. To exploit this, we introduce DeepInsert, a method validated through extensive experiments across different modalities. Our results demonstrate that DeepInsert achieves (a) substantial efficiency gains with negligible impact on performance up to a certain threshold, and (b) a favorable performance-efficiency trade-off when pushed further. The most compelling advantage of DeepInsert is its versatility: it can be seamlessly applied to off-the-shelf MLLMs without any additional hyperparameter-tuning or adaptation.

Our findings open several research directions. First, we must explore the empirical limits of DeepInsert across model sizes and dataset diversity. A natural extension is to video MLLMs, where long-form content poses substantial computational challenges. Second, rigorously characterizing why different models yield varying performance-efficiency trade-offs could advance both our mechanistic understanding of multimodal processing and practical methods for composing pretrained models. Here, we point the reader to recent concurrent work by \citet{hartmanJCBV2025}, that studies this question for VLMs with careful theoretical rigor. Finally, formalizing our hypothesis that multimodal tokens function as queries (see Appendix~\ref{sec:discuss_redundancy}) may illuminate multimodal mechanisms for factual retrieval analogous to those in LLMs.

\section*{Limitations}
\label{sec:limitations}

We only tested our approach on a selective subset of MLLMs. Secondly, despite the offered heuristic, we are unable to offer concrete guarantees on performance without retraining the model, which does expend compute. Finally, since we stick to the baseline evaluation setups, we do inherit their natural deficiencies (for instance, the known drawbacks of captioning performance metrics in MolCA). 

Further, since our work focuses on (pretrained) MLLMs, we suffer from the standard pitfalls associated with such models, including hallucination, equitable access, lack of accountability, and data attribution. At the same time, our proposed method enhances efficiency, potentially lowering computational costs and energy requirements for multimodal model development while maintaining performance, leading to positive societal, economic, and environmental impacts.

\bibliography{references}
\newpage
\appendix

\clearpage

\section{Discussion on Potential Mechanisms Behind Redundancy}
\label{sec:discuss_redundancy}

Herein, we speculate on possible reasons for early multimodal redundancy. The first hypothesis comes from works on mechanistic interpretability, which suggests that the forward pass of an LLM can be broadly delineated into phases: a) the function determination phase that processes the prompt to determine the task that needs to be executed (for instance, \textit{get\_capital()}) and b) the execution of the function for the given input query (for instance, \textit{get\_capital(France) = Paris}) \citep{hendelGG2023, lvCZWLWXY2024, merulloEP2024}. Based on this, we can question whether a similar delineation holds for LLM-based MLLMs: specifically, if we interpret the multimodal tokens as the input query, then the  analogous query execution should take place in the latter layers, which is consistent with our observation that the tokens are used only in the middle layers. Note that query execution here would take an inherently different form, since multimodal tokens are generally either prepended, or inserted in the middle of language prompt. 

Another, more theoretically inclined intuition, comes from recent work in associative memories. Associative memory models for understanding retrieval in transformers have gained traction in recent years, in part due to the strong similarity of the respective inner-product attention mechanisms \cite{ramsauer2021hopfieldnetworksneed, smartBS2025}. Specifically, \citet{Betteti_2025} propose a theoretical model for input-driven dynamics for robust retrieval, wherein the retrieval matrix evolves as a function of an additional input, besides the query. Thus, we can extending our previous analogy, by visualizing the instruction part of the prompt as coaxing the model into the space of vision language retrieval. This coaxing takes some initial processing, beyond which the vision query actually comes into effect. Devising a theoretical retrieval model that can model this 2-stage process can help shed light on any potential benefits that arise from these dynamics (along with the observed redundancy).    

\section{Additional Experiments: Interpretability}

\subsection{Vision-Language Alignment in Pretained Unimodal Models}
\label{sec:VL_alignment}

We begin with some formal exposition to characterize the notion of alignment, borrowed from \citet{huhCWI2024}. A representation is a function $ \mathbf{f} : \mathcal{X} \to \mathbb{R}^n $ that assigns a feature vector to each input in some data domain $\mathcal{X}$. A kernel, $ \mathbf{K} : \mathcal{X} \times \mathcal{X} \to \mathbb{R} $, characterizes how its corresponding representation measures distance/similarity. It is defined as $\mathbf{K}(x_i, x_j) = \langle \mathbf{f}(x_i), \mathbf{f}(x_j) \rangle,$ where $\langle \cdot , \cdot \rangle$ denotes the inner product, $ x_i, x_j \in \mathcal{X} $, and $ \mathbf{K} \in \mathcal{K} $. Finally, a kernel-alignment metric, $ \mathbf{m} : \mathcal{K} \times \mathcal{K} \to \mathbb{R} $, measures the similarity between two kernels, that is, how similar is the distance measure induced by one representation to the distance measure induced by another. As in \citet{huhCWI2024}, we rely on a mutual nearest-neighbor metric that measures the mean intersection of the $k$-nearest neighbor sets induced by two kernels (normalized by $k$).

In essence, we measure the distance between the kernels induced by the representation spaces of language and vision. First, these representations are obtained by passing paired image-caption data through the respective pretrained models. Then, we rely on a mutual $k$-nearest neighbor alignment metric to estimate the alignment between the corresponding feature spaces. We want to understand how alignment varies as a function of model depth, with the intuition that layers capturing ``world semantics" are likely to align better, even across modalities. For the LLaVA v1.5-7b model, its starting components include the vision encoder, clip-vit-large-patch14-336 \cite{radfordKHKS2021}, and the decoder-only LLM vicuna-v1.5-7b model \cite{zhengCSZWS2023}. To compute the alignment metric, we use $k=10$ nearest neighbors over 1024 samples from WIT (Wikipedia-based Image Text; \cite{srinivasanRCBN2021}); additional details can be found in \citet{huhCWI2024}.

\begin{figure*}[htbp]
     \centering 
         \includegraphics[width=1\linewidth]{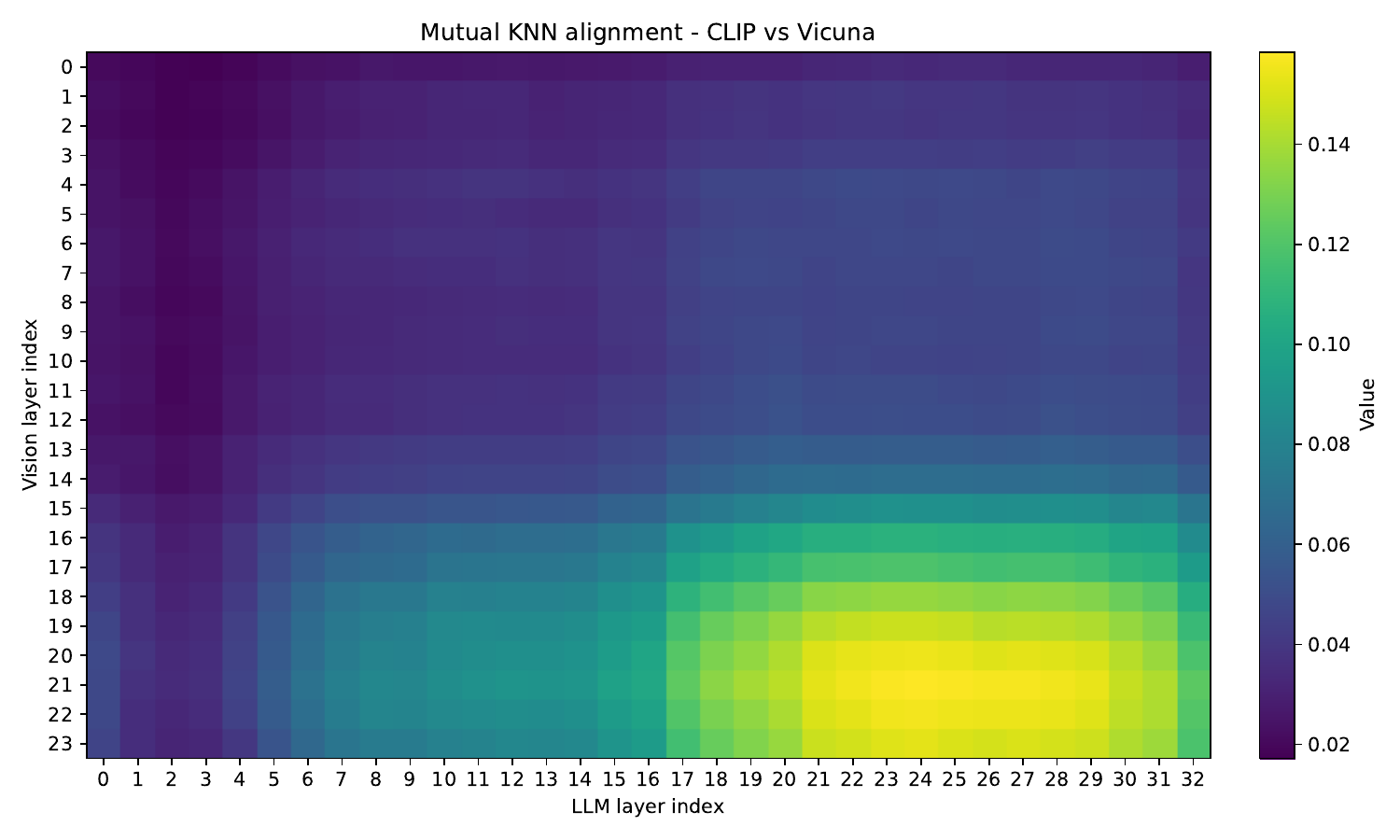}
         \caption{Visualization of intrinsic alignment between pretrained vision (CLIP-based ViT) and language models (Vicuna v1.5), as a function of depth. The vertical axis represents the LLM layer indices, whereas the horizontal axis does the same for the vision encoder. The key point is that for the representations from the latter vision layers (which are generally used as embeddings), alignment becomes stronger as one moves deeper into the LLM.   
         }
        \label{fig:alignment_main}
\end{figure*}

\begin{figure*}[!htbp]
     \centering 
         \includegraphics[width=1\linewidth]{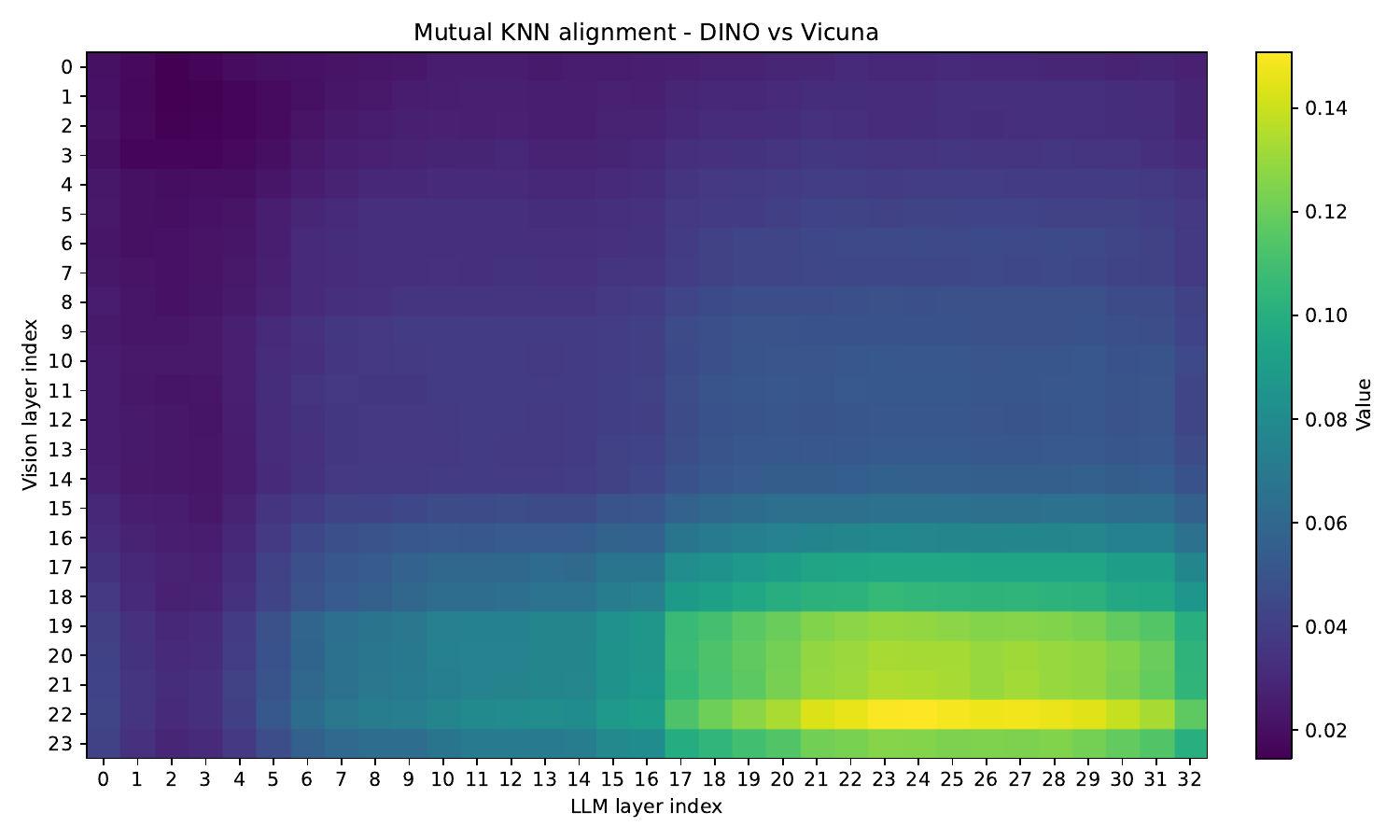}
         \caption{Visualization of intrinsic alignment between pretrained vision (DINO-based ViT) and language models (Vicuna v1.5), as a function of depth. The trend is similar to that for CLIP-based models, in for the representations from the latter vision layers (generally used as embeddings), alignment becomes stronger as one moves deeper into the LLM.   
         }
        \label{fig:alignment_dino}
\end{figure*}

The alignment plot in Fig.~\ref{fig:alignment_main} is in line with our expectations: the penultimate layer of the vision model aligns more with the latter layers of the LLM. This may point to some shared semantics in the deeper layers of the respective models, and can perhaps be exploited in designing alignment frameworks for VLMs. Note that this alignment is more general. Fig.~\ref{fig:alignment_dino} shows a similar trend in DINO-based encoders \cite{caronTMJMBJ2021}. This is remarkable as it points to a more general preexisting alignment, since DINO vision encoders are---unlike CLIP---trained in a self-supervised language-agnostic fashion.

\clearpage

\subsection{Where Visual Tokens Matter}
\label{sec:WVM_rem}
\begin{figure*}[!hbtp]
     \centering 
         \includegraphics[width=0.8\linewidth]{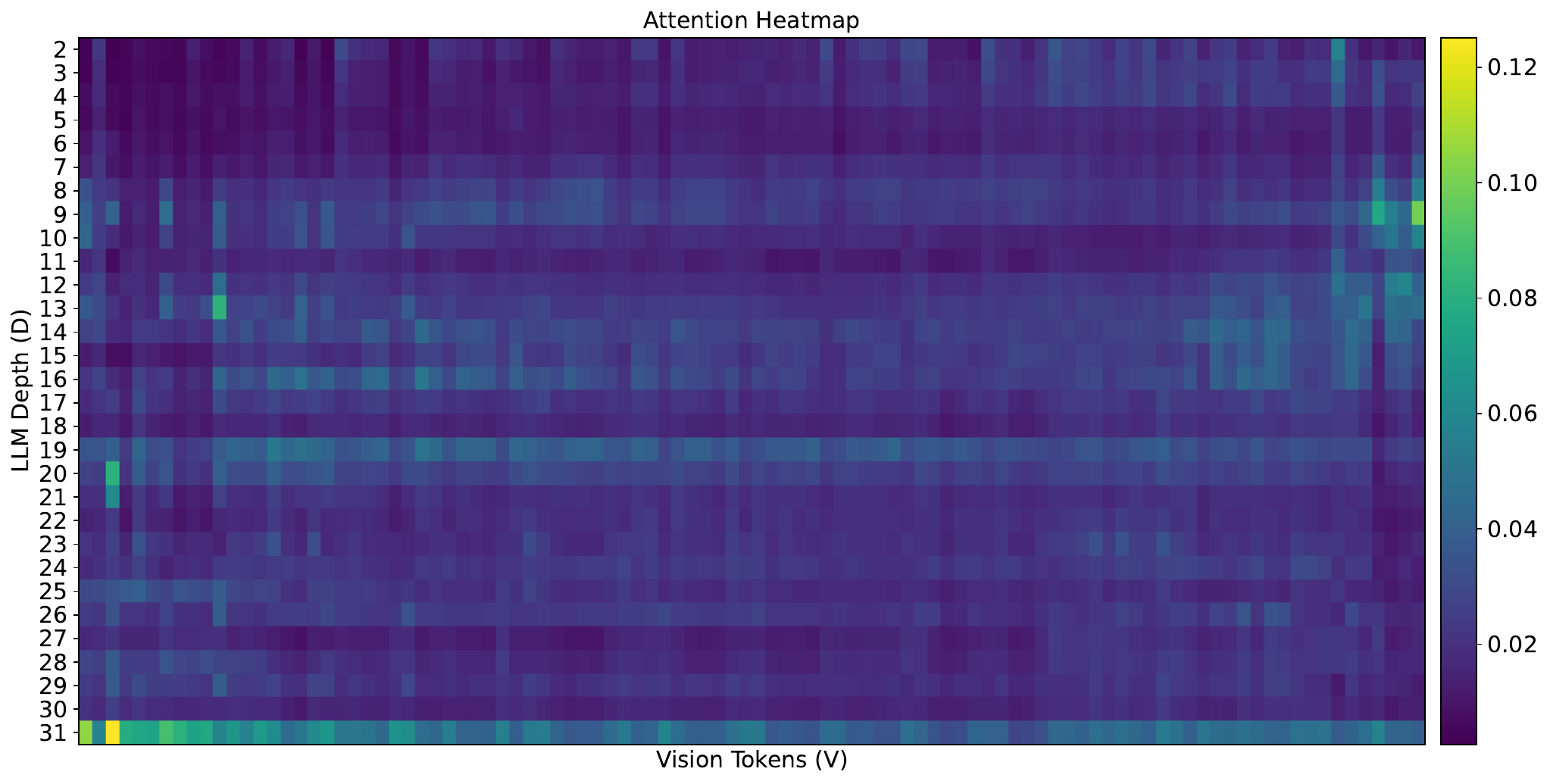}
         \caption{Illustration of attention activity, averaged over all heads.
         }
        \label{fig:attention_viz_all_heads}
\end{figure*}

\begin{figure*}[!ht]
     \centering 
         \includegraphics[width=0.8\linewidth]{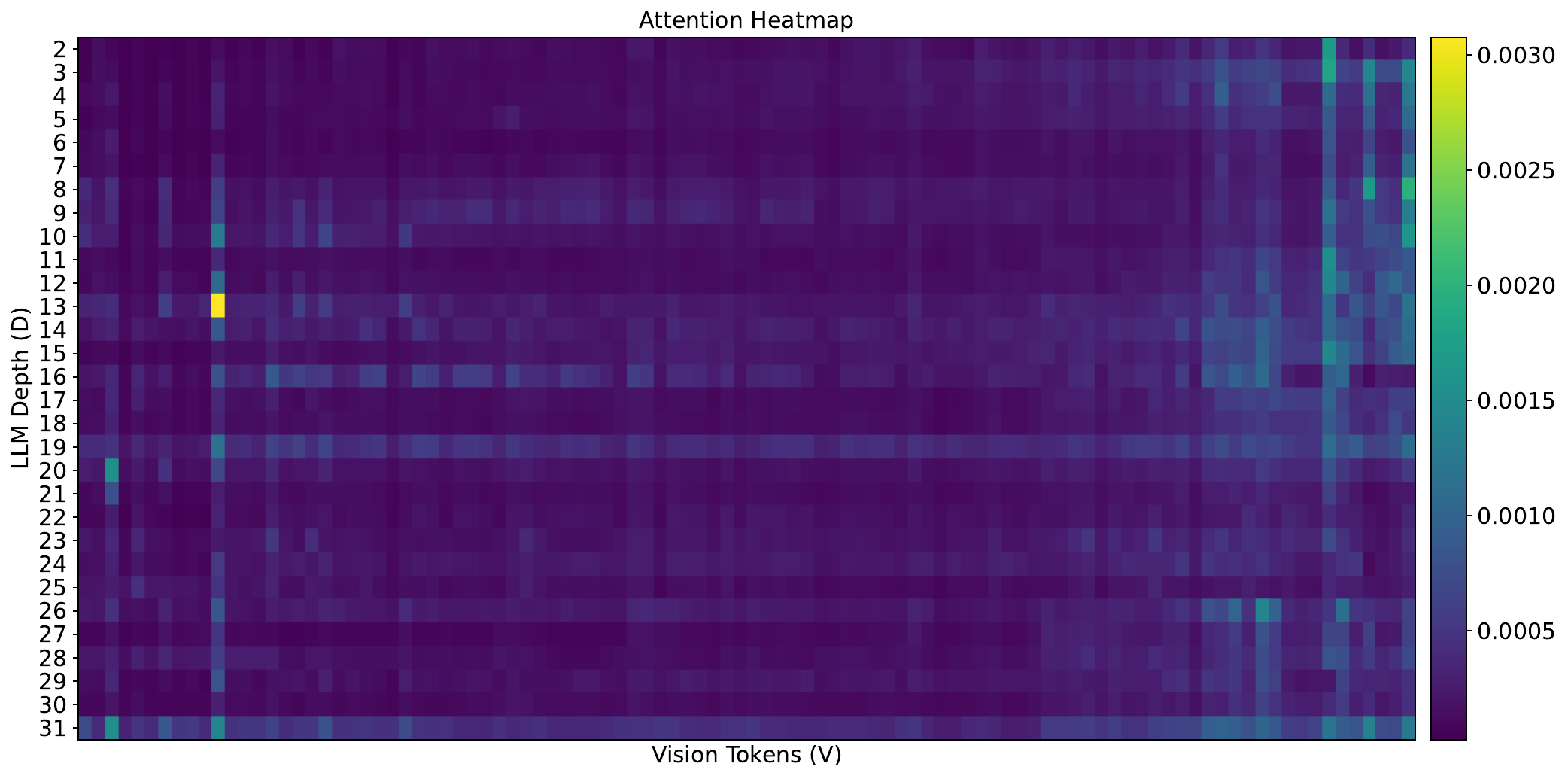}
         \caption{Illustration of attention activity, without token specific normalization.
         }
        \label{fig:attention_viz_no_normalized}
\end{figure*}

Here, we present the complete list of visualizations, to comprehensively demonstrate the middle layer activation. We begin by visualizing the attention activation for all layers. As we can see, the activation in the middle, while still prominent, is overshadowed by the flurry in the first two layers. Nevertheless, as mentioned in the main paper, this is likely of little import for analyzing the general trend. We also demonstrate that the visualization when done over the average behavior of activation heads as opposed to top-5 (see Fig. \ref{fig:attention_viz_all_heads}), or without normalization (see Fig. \ref{fig:attention_viz_no_normalized}) does not make a difference in terms of the nature of our conclusions. For the latter, even though we see the high magnitude CLIP tokens dominate the attention matrix, their activity is also more middle layers heavy.
\begin{figure*}[!b]
     \centering 
         \includegraphics[width=0.8\linewidth]{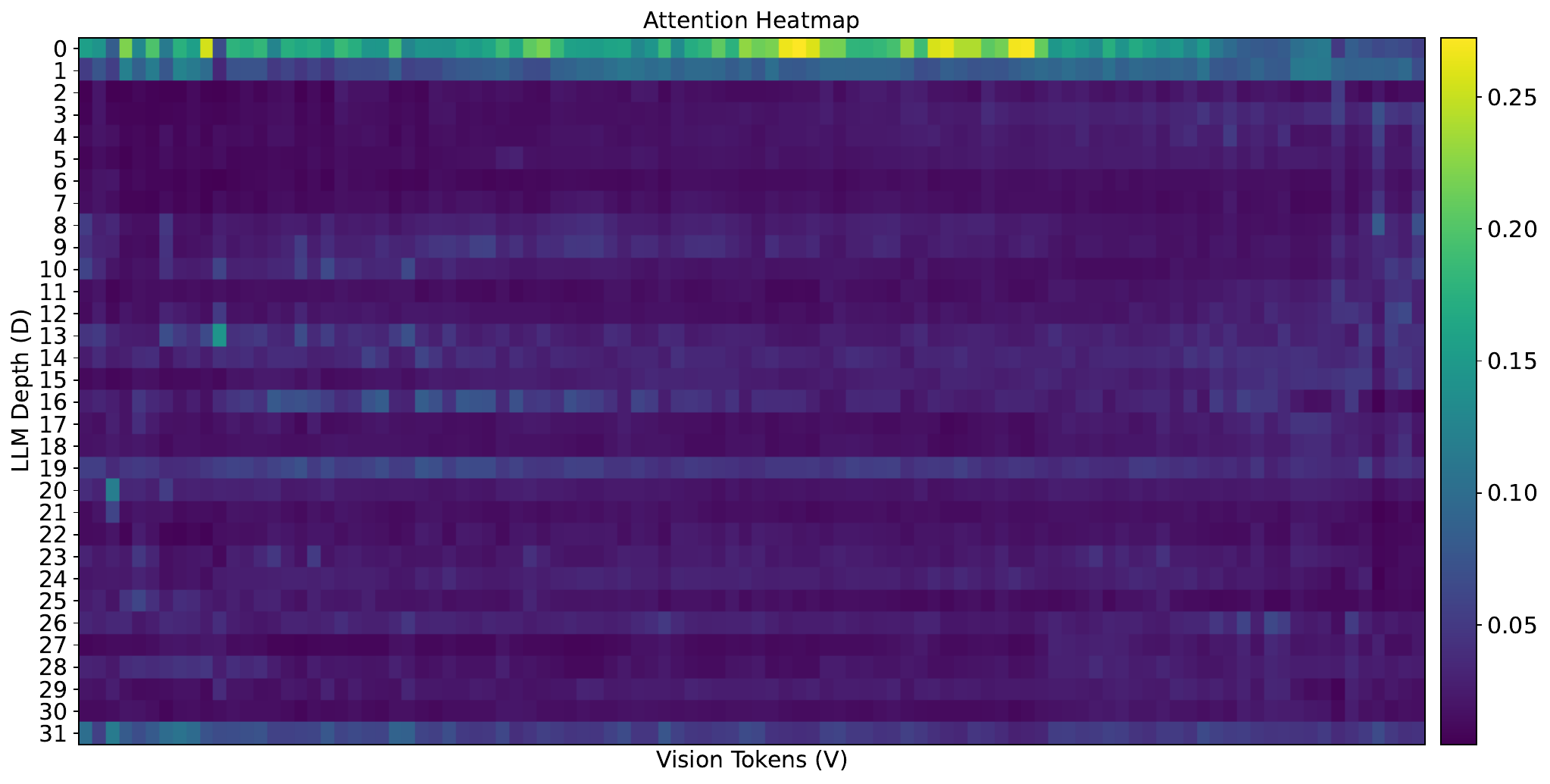}
         \caption{Full illustration of vision to language token activations (including the first two layers), in the forward pass of a LLaVA-v1.5-7B model.    
         }
        \label{fig:attention_viz_avg_all_layers}
\end{figure*}

\clearpage

\section{Layer Selection Heuristic}
\label{sec:layer_heuristic}

The key idea is as follows: if there is intrinsic evidence that multimodal tokens are not being used in the early layers, can we skip them even in baseline pretrained models? Our somewhat surprising find is that it is possible to do so, without any finetuning. Specifically, we load the pretrained model weights of the MLLM (LLaVA-7B or LTU-7B) onto the DeepInsert variant, by simply letting the multimodal adapter insert tokens into the latter layer.  

\begin{table}[h]
\centering
\begin{minipage}{0.45\columnwidth}
\centering
\resizebox{0.75\columnwidth}{!}{
\begin{tabular}{@{}lc@{}}
\toprule
\textbf{Model} & \textbf{POPE} \\ \midrule
LLaVA  & 86.0 \\
DI-4 & 84.4 \\
DI-8  & 77.9 \\
DI-12 & 56.3 \\
DI-16 & 47.8 \\
\bottomrule
\end{tabular}}
\subcaption{LLaVA on POPE}
\label{tab:llava_untrained}
\end{minipage}
\hfill
\begin{minipage}{0.5\columnwidth}
\centering
\resizebox{\columnwidth}{!}{
\begin{tabular}{@{}lcc@{}}
\toprule
\textbf{Model} & \textbf{ESC50} & \textbf{VS} \\ \midrule
LTU & 81.30 & 56.39 \\
DI-4 & 81.70 & 56.36 \\
DI-8  & 74.50 & 41.21 \\
DI-12 & 66.40 & 31.05 \\
DI-16 & 32.35 & 32.36 \\
\bottomrule
\end{tabular}}
\subcaption{LTU on ESC50 and VS}
\label{tab:ltu_untrained}
\end{minipage}
\vspace{0.2cm}
\caption{Performance across layers for baseline pretrained weights on vision (left) and audio (right) benchmarks.}
\label{tab:untrained_combined}
\end{table}



From Table \ref{tab:llava_untrained} and \ref{tab:ltu_untrained}, we note that both models retain competency even when layers are skipped. While the performance drops in LLaVA are modest yet noticeable, LTU surprisingly improves for DI-4, suggesting a strong implicit redundancy, at least for classification benchmarks. We also note that these trends are also indicative of which model/modality are likely to be more susceptible to DeepInsert. 

We did however notice that with both models, the lengths of text generation seems to monotonically decrease (to the point that the evaluation benchmark for captioning seems to fail for LTU). Therefore, direct insertion without retraining degrades performance on generative benchmarks and there is still a need to retrain models from scratch.  

\section{Predicting Insert Layer with RL}
\label{app:rl_layer_selection}

We consider a pretrained and frozen LLaVA model augmented with a lightweight adapter whose goal is to determine the transformer layer at which multimodal embeddings should be integrated. The adapter takes prompt embeddings as input and outputs a discrete prediction over layers, corresponding to the insertion point for multimodal information.

The adapter is trained using reinforcement learning (RL) with a reward function designed to balance task performance and computational efficiency. Concretely, the reward consists of two components: (i) the negative next-token prediction loss on the given datapoint, which encourages performance retention, and (ii) a redundancy reward that incentivizes skipping later layers. The redundancy reward is normalized to the range $[0,1]$, where $0$ indicates no layer skipping and $1$ corresponds to skipping all layers. These two terms are combined via a weighting coefficient $\lambda$, which controls the trade-off between performance and efficiency. By adjusting $\lambda$, the optimization can be biased toward either higher accuracy or reduced computation. 

\begin{figure}[t]
    \centering
    \includegraphics[width=0.9\linewidth]{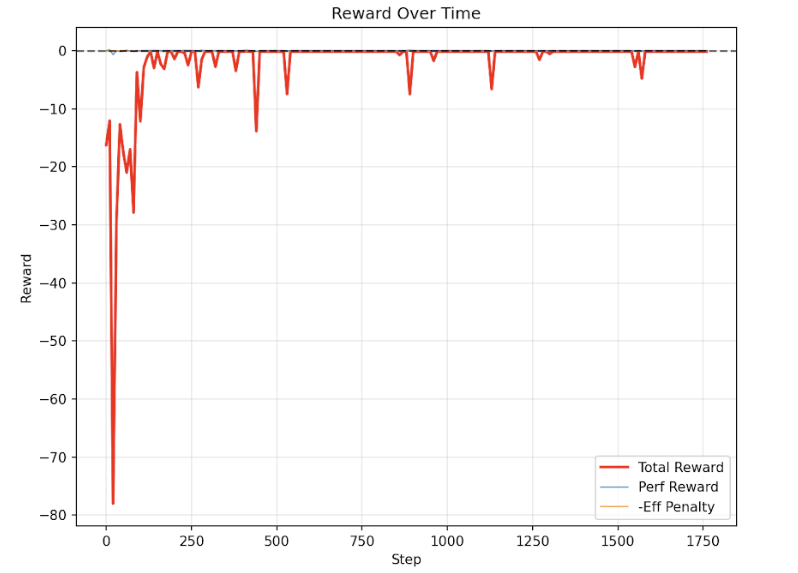}
    \caption{
   Total reward (red) over optimization steps, decomposed into the performance reward (blue) and the efficiency-related penalty (orange). The dashed horizontal line indicates zero reward.
    }
    \label{fig:reward_over_time}
\end{figure}

The adapter is trained on a small subset (10\%) of the LLaVA instruction fine-tuning dataset, consisting of samples from COCO, TextVQA, and GQA. The adapter architecture is a simple multilayer perceptron (MLP). 

The Figure ~\ref{fig:reward_over_time} illustrates the evolution of the reward over time. In the early stages, the total reward exhibits large negative values and high variance, reflecting unstable exploration and frequent violations of efficiency constraints. As training progresses, the policy rapidly improves, with the total reward increasing toward zero and fluctuations becoming less severe. After convergence, the reward remains close to zero for most steps, punctuated by occasional negative spikes. Overall, the trend indicates successful learning of a policy that balances task performance with efficiency considerations, achieving stable behavior while respecting the imposed penalty structure.

The average predicted insertion depth is approximately $0.14$ in normalized layer index, corresponding to layer $4.8$ in a 32-layer transformer (as used in 7B-scale models). This closely aligns with our empirical findings for DI-4.

While further exploration is necessary, these results suggest a principled, data-driven approach for identifying suitable insertion layers for multimodal integration.


\section{Integrating with token-reduction techniques}
\label{sec:token_reduction-integration}

As we mention in Sec.~\ref{sec:related_vlm}, DeepInsert in complementary to token-reduction/token-pruning techniques such as FastV~\cite{chen2024fastv}, VTW~\cite{lin2024vtw}, and PruMerge~\cite{shangCXLY2024}, and applying these to DeepInsert models can give further efficiency gains. These techniques drop less important tokens in the attention mechanism, decreasing the computational load of the model. Specifically, we find that DeepInsert models generally respond better to token-pruning methods, compared to the baseline. In line with recommended hyperparameters, for FastV, we use K=3 (layer to start pruning) and R=25\% (retention ratio). For DeepInsert, K=3 implies pruning 3 layers after the insertion layer. Similarly, we use K=16 for VTW (implying tokens exit after layer 16). For PruMerge, we use the training free-variant with R=12.5\%. Table~\ref{tab:token-reduction} lists the average performance over AI2D, POPE, and MMMU, when applying the token-reduction techniques, to the baseline LLaVA-7B model as well as the DeepInsert variants, highlighting that the latter is equally or more robust to pruning methods.

\begin{table}[h]
\centering
\resizebox{\columnwidth}{!}{
\begin{tabular}{@{}lcccc@{}}
\toprule
\textbf{Model}  & \textbf{Baeline Acc.}         & \textbf{+FastV}  & \textbf{+VTW} & \textbf{+PruMerge} \\ \midrule
LLaVA & 59.4                      & 57.8 (-1.6)	        &  59.9 (+0.5)  & 59.8(+0.4)  \\
DI-4            & 59.4                      & 59.2 (-0.2)	        &  59.9 (+0.5)   & 59.8(+0.4) \\ 
DI-8            & 57.6                      & 58.0 (+0.4)	        &  58.0 (+0.4)    & 58.1(+0.5)\\
DI-12           & 57.3                      & 57.9 (+0.6)	        &  57.9 (+0.6)    & 57.8(+0.5) \\
\bottomrule
\end{tabular}}
\vspace{0.4cm}
\caption{Average performance of LLaVA-7B and DeepInsert models, showing that DeepInsert models are equally or more robust to FastV, VTW, and PruMerge.}
\label{tab:token-reduction}
\end{table}


\section{Additional Experiments: MLLMs}

\subsection{BLIP}
\label{subsec:BLIP}

BLIP (Bootstrapped Language-Image Pretraining)  is a pioneering approach for vision-language modeling \cite{liLXH2022}. It introduces a lightweight Querying Transformer (QFormer) that extracts information from the embeddings of the vision encoder. Vision information is transferred into a set of learnable query tokens (32 embeddings of dimension 768) via the QFormer's cross-attention layers, which are then fed into the LLM along with the language prompt. Further improvements by: a) scaling the LLM and pretraining dataset \cite{JunnanLSH2023}, and b) multi-task instruction finetuning \cite{daiLLTZWLFH2023} allows the model to solve a variety of tasks like image captioning, visual question answering, and multimodal reasoning. 

In our experiment, we will consider one such model to demonstrate the potential of DeepInsert. Specifically, we consider the VLM's construction with FlanT5-base \cite{chungHLZT_et_al_2022} (encoder-decoder architecture, 350M parameter model) as our LLM and the Eva-clip-g/14 \cite{sunFWWC2023}, a CLIP-based vision transformer, for our frozen image encoder. The QFormer follows the default architecture and initialization as in \citet{daiLLTZWLFH2023}. Since FlanT5-base has an encoder-decoder architecture, insertion bypasses the encoder entirely. Instead of feeding the Qformer outputs (query tokens) as inputs to the encoder, we feed them directly into the decoder by prepending them to the language tokens. The choice of bypassing the encoder is intuitive, since the encoder output space is likely to capture the semantics of the language prompt and therefore align better with the image semantics captured in the QFormer tokens.

Note that we do not intend to follow the training recipe for the InstructBLIP models. Indeed, it is impossible to replicate the model for two reasons: a) lack of information on training recipe, including learning rate schedules and dataset sampling ratios, and b) parts of the LAION dataset, the key component for pretraining, was taken offline so as to comply with U.S.\ federal law, cf.~\citet{thiel2023}. Furthermore, the high-level recipe calls for multi-epoch pretraining on a dataset with roughly 130M image-text pairs, which are beyond the scope of this work. Instead, we define a smaller setup, which still allows us fair testing of DeepInsert but with a reasonable compute budget. Recall that the end-to-end training phase for InstructBLIP involves two steps: the pretraining step with image-caption pairs, followed by the multi-task instruction finetuning step.
Consequently, we structure the experiment in a similar vein, by first pretraining efficiency, followed by a comparison post multi-task instruction finetuning.  

\subsubsection{Pretraining Efficiency}

\begin{figure}[!htb]
     \centering
         \includegraphics[width=0.9\linewidth]{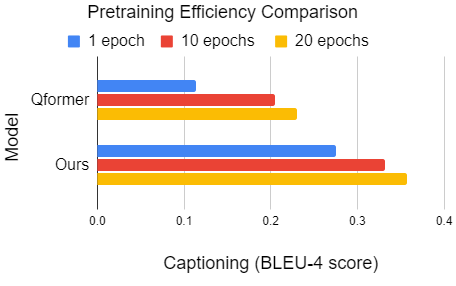}
         \caption{Pretraining comparison for Baseline (QFormer) vs. DeepInsert (Ours) pipelines. Higher score indicates that DeepInsert achieves better alignment, and in much fewer epochs.}
        \label{fig:pretrain}
\end{figure}

Reflecting the InstructBLIP setup, we consider image-caption pretraining with the COCO captions dataset \cite{chenFLVGDZ2015} as the first stage. The goal of this experiment is to compare the level of alignment between the QFormer and the LLM, as training proceeds. Specifically, we measure the best possible captioning score achieved within a fixed number of epochs (1, 10 and 20) for each framework. This is done in accordance with different hyperparameter and learning rate configurations available in the BLIP repository and for each variant, we report the highest performance across configurations. As shown in Fig.~\ref{fig:pretrain}, we find that our framework not only achieves better peak performance, but also does it in much fewer epochs. We limit our report to 20 epochs as both models achieve over 95\% of their respective peak scores within that frame (the peak is achieved at around 50 epochs). 

\subsubsection{Multi-task Training}

Next, we consider instruction finetuning for two tasks: Captioning and Visual Question Answering (VQA).
For each task, we sample from a fixed set of text prompts in every training iteration. 

To emulate InstructBLIP's training stages, we first pretrain with image-caption pairs for 20 epochs, and then perform multi-task instruction finetuning for captioning and VQA  for 15 combined epochs (uniformly sampling from the datasets without replacement).  For captioning, we consider the COCO captions dataset as before and for visual question answering, we consider VQAv2 \cite{goyalKSBP2017} for training. Instruction finetuned VLMs can demontrate remarkable zero-shot abilities on unseen datasets, and so we also consider OKVQA (Outside Knowledge VQA) \cite{marinoRFM2019} for zero-shot evaluation. Coco Captions and VQAv2 share the same train-val-test image split that aims to prevent train-test information leakage for a fair evaluation. 

Results are given in Table \ref{tab:blip_performance}. We find that for the designated tasks, our method achieves significant improvements over the standard QFormer.  We also see a remarkable gap in zero-shot OKVQA performance, with the grounded model even outperforming the reported performance of some of the largest BLIP-2 models \cite{JunnanLSH2023}. (InstructBLIP includes OKVQA during training, so no direct comparison is possible.) 

\begin{table}[!htbp]
\begin{center}
\resizebox{\columnwidth}{!}{
\begin{tabular} {@{}lcccc@{}}
 \toprule
 \textbf{Model} & \textbf{COCO Cap. (BLEU-4)} & \textbf{VQAv2 (\%)}  & \textbf{OKVQA (0-shot) (\%)} \\
 \midrule
 \multicolumn{4}{l}{} \\
 Baseline  & 20.9 & 55.4 & 28.8  \\
 DI (Ours)   & $\bm{36.2}$ & $\bm{66.8}$ & $\bm{39.0}$ \\
\hline
\end{tabular}}
\caption{Performance comparison for BLIP-inspired multi-task instruction finetuned models, including 0-shot evaluation on unseen dataset. We find that in our custom setup, bypassing the encoder and inserting directly to the decoder can yield significant performance gains. }
\label{tab:blip_performance}
\end{center}
\end{table}

\subsection{Open-LLaVA-NeXT}
\label{subsec:llava_next}

Herein, we extend DeepInsert to a the newer LLaVA-Next-7B (the open reproduction) \cite{liuLLLZSL2024, chenX2024}, with a much larger vision context yielding performance improvements. Below, we report the average performances in Table \ref{tab:llava_next}, indicating the robustness of DeepInsert to exploit early layer redundancy even when number of multimodal tokens are much higher ($\sim$2k). We note here that unlike retraining the baseline models in the main paper, we directly used the checkpoint from Huggingface for the baseline due to limited compute, meaning the DeepInsert models had about 1-2k missing data points, which explains the slight performance drop.

\begin{table}[h]
\centering
\resizebox{\columnwidth}{!}{
\begin{tabular}{@{}lcccccc@{}}
\toprule
\textbf{Model} & \textbf{AI2D} & \textbf{GQA} & \textbf{POPE} & \textbf{MMMU} & \textbf{TextVQA (lite)} & \textbf{Avg.} \\ \midrule
LLaVA-NeXT & 64.6 & 64.2 & 87.3 & 39.0 & 63.4 & 63.7 \\
DI-4 & 63.7 & 63.8 & 87.4 & 36.9 & 64.3 & 63.2 \\
DI-6 & 62.7 & 63.6 & 87.0 & 37.1 & 61.9 & 62.4 \\ 
\bottomrule
\end{tabular}}
\vspace{0.4cm}
\caption{Performance of DeepInsert on Open-LLaVA-NeXT-7B.}
\label{tab:llava_next}
\end{table}

\subsection{Training Loss Comparison on LLaVA}

\begin{figure}[thbp!]
     \centering 
         \includegraphics[width=\columnwidth]{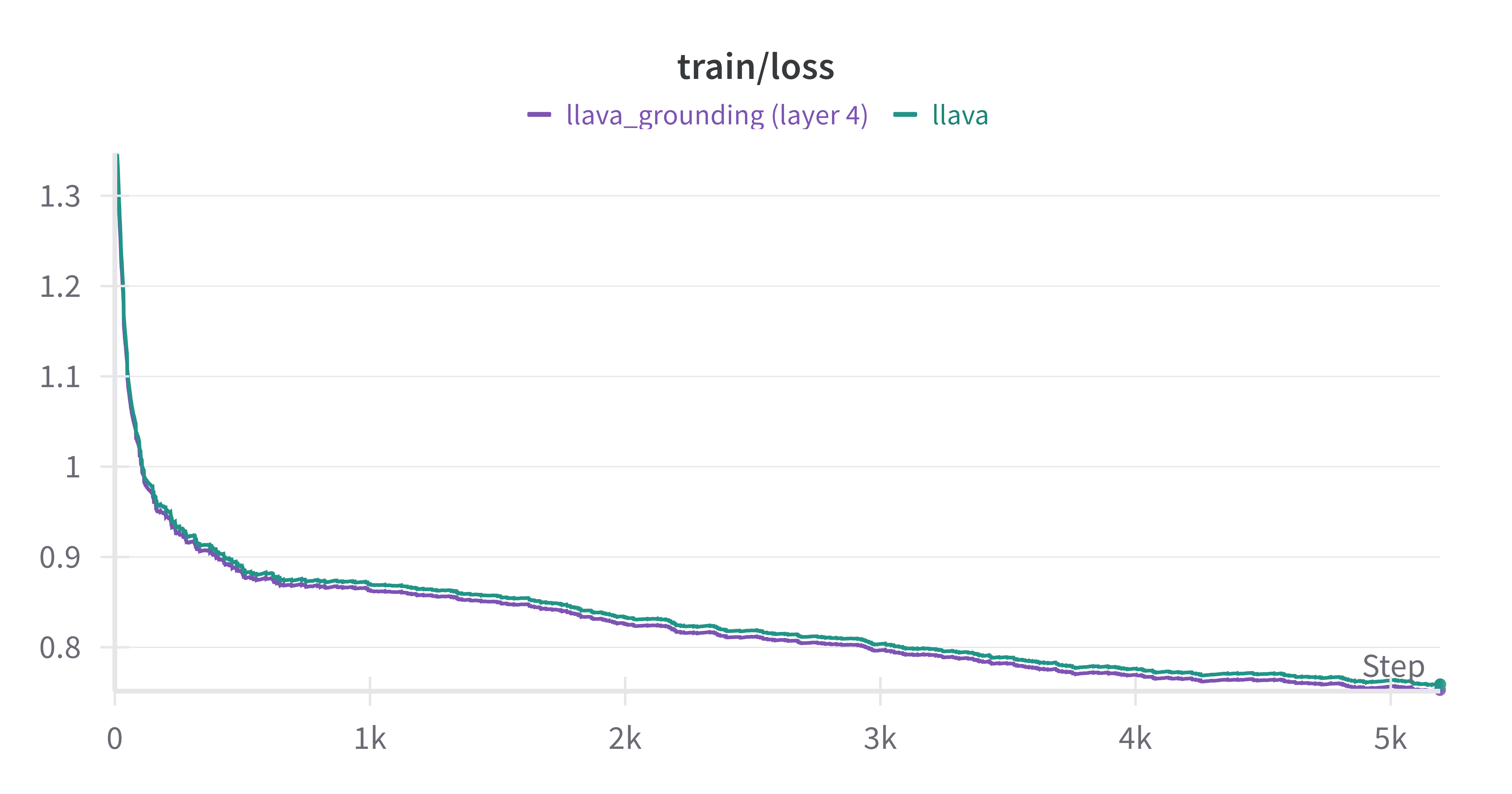}
         \caption{Comparison of training losses of LLaVA finetuning stage, for the baseline model against best performing grounded model (layer-4), revealing a tendency to achieve lower next token prediction loss and potentially better learning.}
        \label{fig:training_losses}
\end{figure}


\section{Efficiency Analysis}
\label{sec:app_efficiency}

The number of FLOPs performed by a transformer model in a forward pass can be analytically
expressed as a function of the transformer's various hyperparameters such as model hidden dimension,
feedforward layer dimension, and number of layers.
In this section, we extend the analysis to DeepInsert and compare the theroetically expected
FLOPs with the empirically measured runtime.
Let the transformer have $N$ layers with a hidden dimension of 
$d_\text{model}$ and a feedforward dimension of $d_\text{ff}$.
Let the number of heads for multihead attention be given by $n_\text{head}$.
Assume the number of multimodal tokens is $L_\text{mm}$ and the number of text tokens
in a certain forward pass is $L_\text{text}$.
Let the DeepInsert layer be given by $N_\text{DI}$.

The total number of FLOPs (additions plus multiplications) in one forward pass can be divided into 
the following parts: 
(i) the projections into Query ($Q$), Key ($K$), and Value ($V$) matrices,
(ii) the self attention mechanism, 
(iii) the feedforward layers, 
and (iv) other computations such as activations, layer norm, etc.
Since (iv) is negligible compared to (i), (ii), and (iii), we ignore it here.

Expanding on (i), the input $X\in\mathbb{R}^{L\times d_\text{model}}$ needs to be multiplied
with projection matrices $W_q,W_k,W_v\in\mathbb{R}^{d_\text{model}\times d_\text{model}}$ to
give $Q,K,V$ respectively.
For an input sequence of length $L$, each $XW$ matrix multiplies require $2Ld_\text{model}^2$ operations including both additions and multiplications, giving us
\begin{align}
    \text{Projection FLOPs} = 6Ld_\text{model}^2.
    \label{eq:projection-flops}
\end{align}

Moving to (ii), we first need to compute the $QK^\top$ matrix, and since $Q$ and $K$ are both in
$\mathbb{R}^{L\times d_\text{model}}$, this requires $n_\text{head}\times2L^2\frac{d_\text{model}}{n_\text{head}}=2L^2d_\text{model}$ FLOPs.
We then apply a row-wise softmax to get the $L\times L$ attention matrix for each head,
which requires another $2L^2n_\text{head}$ FLOPs.
Next, the $L\times L$ attention matrices are multiplied with the value matrices which takes up another
$2L^2d_\text{model}$ FLOPs across all heads.
Finally, the output projection takes up another $2Ld_\text{model}^2$ FLOPs.
Adding all these up, and using the approximation $n_\text{head}<<d_\text{model}$, we have
\begin{align}
    \text{Attention FLOPs} = 4L^2d_\text{model} + 2Ld_\text{model}^2.
    \label{eq:attention-flops}
\end{align}

Computing (iii) is more straightforward as the feedforward layers are pairs of linear layers, with
dimensions $d_\text{model}\times d_\text{ff}$ and $d_\text{ff}\times d_\text{model}$.
This gives us
\begin{align}
    \text{Feed forward FLOPs} = 4Ld_\text{model}d_\text{ff}.
    \label{eq:ff-flops}
\end{align}

Adding \eqref{eq:projection-flops}, \eqref{eq:attention-flops}, and \eqref{eq:ff-flops} gives us our final analytical expression for the number of FLOPs per layer when the sequence length
is $L$:
\begin{align}
    \text{FLOPs per layer for $L$ tokens}& \nonumber\\
    = 8Ld_\text{model}^2 + 4L^2d_\text{model} + &4Ld_\text{model}d_\text{ff}.
    \label{eq:flops-per-layer}
\end{align}

Using DeepInsert, the first $N_\text{DI}$ layers only see $L_\text{text}$ tokens, while
the remaining $N-N_\text{DI}$ layers see all the $L_\text{mm}+L_\text{text}$ tokens.
This give us the total number of FLOPs for deep insertion at layer $N_\text{DI}$:
\begin{align}
&\text{FLOPs}(N_\text{DI}) \nonumber\\
&= 
N\Big(8(L_\text{text}+L_\text{mm})d_\text{model}^2 \nonumber\\
&\qquad\qquad+ 4(L_\text{text}+L_\text{mm})^2d_\text{model} \nonumber\\
&\qquad\qquad\qquad\qquad+ 4(L_\text{text}+L_\text{mm})d_\text{model}d_\text{ff}\Big) \nonumber\\
&-N_\text{DI}\Big(8L_\text{mm}d_\text{model}^2 \nonumber\\
&\qquad\qquad+4(2L_\text{text}+L_\text{mm})L_\text{mm}d_\text{model}\nonumber\\
&\qquad\qquad\qquad\qquad+4L_\text{mm}d_\text{model}d_\text{ff}\Big).
\label{eq:deepinsert-flops}
\end{align}

As we see from \eqref{eq:deepinsert-flops}, the number of FLOPs decreases monotonically as a function
of $N_\text{DI}$.
Further, a larger number of multimodal tokens $L_\text{mm}$ leads to a steeper decrease in FLOPs as
we insert deeper into the network.

\begin{figure}[t]
    \centering
    \includegraphics[width=\columnwidth]{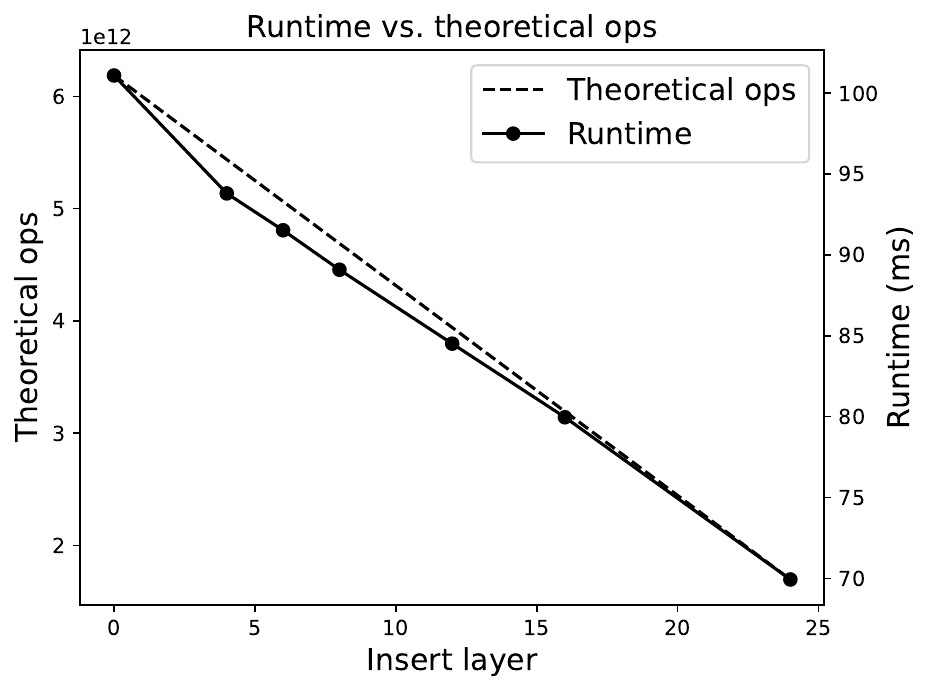}
    \caption{Comparing the model runtime with the theoretically estimated FLOPs from \eqref{eq:deepinsert-flops}. As we see, both curves match very closely.}
    \label{fig:deepinsert-flops-runtime}
\end{figure}

To validate \eqref{eq:deepinsert-flops}, we empirically measure the runtime of LLaVA on 80 GB A100 GPUs and plot the both the runtime and the theoretically estimated FLOPs against the insertion layer in
Fig.~\ref{fig:deepinsert-flops-runtime}.
As we see in Fig.~\ref{fig:deepinsert-flops-runtime}, there is a very close match between the theoretically estimated FLOPs of \eqref{eq:deepinsert-flops} and the empirically measured runtime
of the forward pass of the model, suggesting that \eqref{eq:deepinsert-flops} can be a useful
proxy for the computational efficiency of the model when insertion is done at various deep layers.

\section{Potential Risks}
Efficiency gains can lower deployment costs, potentially enabling wider, sometimes unguarded use. If models hallucinate or provide ungrounded rationales, this may scale harmful outcomes (e.g., disinformation, low-friction generation of misleading content) when safeguards are absent. As with other MLLMs, the system remains vulnerable to adversarial prompts and distribution shifts. Efficiency itself neither increases nor decreases susceptibility, but wider, cheaper access can increase attack surface (e.g., more attempts, faster iteration by adversaries).

\section{AI Usage}
AI models (part of Overleaf native tools) were used in improving the writing of the paper.


\clearpage

\end{document}